\documentclass[10pt,twocolumn,letterpaper]{article}

\usepackage{cvpr}              % To produce the CAMERA-READY version
\usepackage{graphicx}
\usepackage{amsmath}
\usepackage{amssymb}
\usepackage{booktabs}

\usepackage[accsupp]{axessibility} % Improves PDF readability for those with disabilities.

\usepackage{multirow}
\usepackage{bbm}
\usepackage{comment}

\usepackage{pifont}% http://ctan.org/pkg/pifont
\newcommand{\cmark}{\ding{51}}%
\usepackage[pagebackref,breaklinks,colorlinks]{hyperref}

\usepackage[capitalize]{cleveref}
\crefname{section}{Sec.}{Secs.}
\Crefname{section}{Section}{Sections}
\Crefname{table}{Table}{Tables}
\crefname{table}{Tab.}{Tabs.}

\def\cvprPaperID{5871} % *** Enter the CVPR Paper ID here
\def\confName{CVPR}
\def\confYear{2022}

\begin{document}

\title{
OVE6D: Object Viewpoint Encoding for Depth-based 6D Object Pose Estimation
}

% \author[1]{Dingding Cai}
% \author[2]{Janne Heikkilä}
% \author[1]{Esa Rahtu}
% \affil[1]{Tampere University} \and \affil[2]{University of Oulu}
% \affil[1]{\textit{\{dingding.cai, esa.rahtu\}@tuni.fi}}
% \affil[2]{\textit{janne.heikkila@oulu.fi}}

% \author{
% Dingding Cai\\
% Tampere University\\
% {\tt\small \email{dingding.cai@tuni.fi}}
% \and
% Janne Heikkilä\\
% University of Oulu\\
% {\tt\small \email{janne.heikkila@oulu.fi}}
% \and
% Esa Rahtu\\
% Tampere University\\
% {\tt\small \email{esa.rahtu@tuni.fi}}
% }

\author{
Dingding Cai$^1$,~~~~~ Janne Heikkilä$^2$,~~~~~ Esa Rahtu$^1$ \\
$^1$Tampere University,~~~~~ $^2$University of Oulu \\
{\tt\small \{dingding.cai, esa.rahtu\}@tuni.fi,  \tt\small janne.heikkila@oulu.fi}
}

\maketitle
%%%%%%%%% ABSTRACT
\begin{abstract}
   This paper proposes a universal framework, called OVE6D, for model-based 6D object pose estimation from a single depth image and a target object mask. Our model is trained using purely synthetic data rendered from ShapeNet, and, unlike most of the existing methods, it generalizes well on new real-world objects without any fine-tuning. We achieve this by decomposing the 6D pose into viewpoint, in-plane rotation around the camera optical axis and translation, and introducing novel lightweight modules for estimating each component in a cascaded manner. The resulting network contains less than 4M parameters while demonstrating excellent performance on the challenging T-LESS and Occluded LINEMOD datasets without any dataset-specific training. We show that OVE6D outperforms some contemporary deep learning-based pose estimation methods specifically trained for individual objects or datasets with real-world training data. 
%   The implementation and the pre-trained model will be made publicly available
 The implementation is available at \href{https://github.com/dingdingcai/OVE6D-pose}{https://github.com/dingdingcai/OVE6D-pose}.
\end{abstract}

% 3,780,741 (3.78M)parameters 

%%%%%%%%% BODY TEXT
\section{Introduction}
\label{sec:intro}

% Please follow the steps outlined below when submitting your manuscript to the IEEE Computer Society Press.
% This style guide now has several important modifications (for example, you are no longer warned against the use of sticky tape to attach your artwork to the paper), so all authors should read this new version.

\begin{figure}[ht]
\centerline{\includegraphics[width=0.90\linewidth]{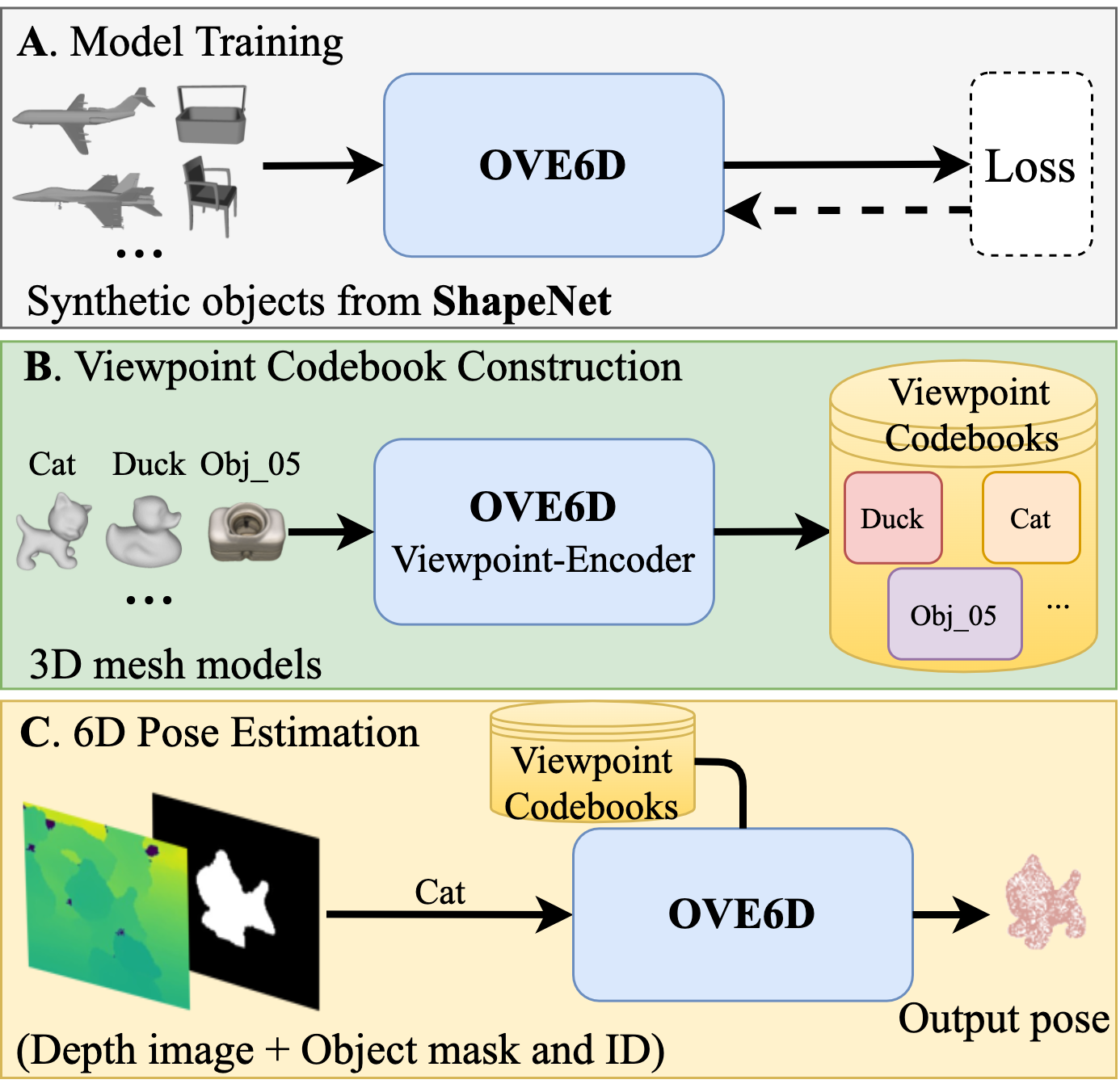}}
\caption{
\textbf{A)} We propose a single universal pose estimation model (called OVE6D) that is trained using more than 19,000 synthetic objects from ShapeNet. \textbf{B)} The pre-trained model is applied to encode the 3D mesh model of the target object (unseen during the training phase) into a viewpoint codebook. \textbf{C)} At the inference time, OVE6D takes a depth image, an object segmentation mask, and an object ID as an input, and estimates the 6D pose of the target object using the corresponding viewpoint codebook. New object can be added by simply encoding the corresponding 3D mesh model and including it into the codebook database (\textbf{B}).  
%given an observed depth image, the target object segmentation mask and its identity, OVE6D estimates 6D pose for the target object. 
%The proposed OVE6D is a single universal model that is trained with more than 19,000 synthetic objects from ShapeNet. \textbf{B)}. The trained OVE6D model is used to construct object viewpoint codebook database for novel real-world objects using 3D mesh models. \textbf{C)}. At the inference time, given an observed depth image, the target object segmentation mask and its identity, OVE6D estimates 6D pose for the target object.
}
\label{fig:intro}
\end{figure}

\begin{figure}[t]
\centerline{\includegraphics[width=0.82\linewidth]{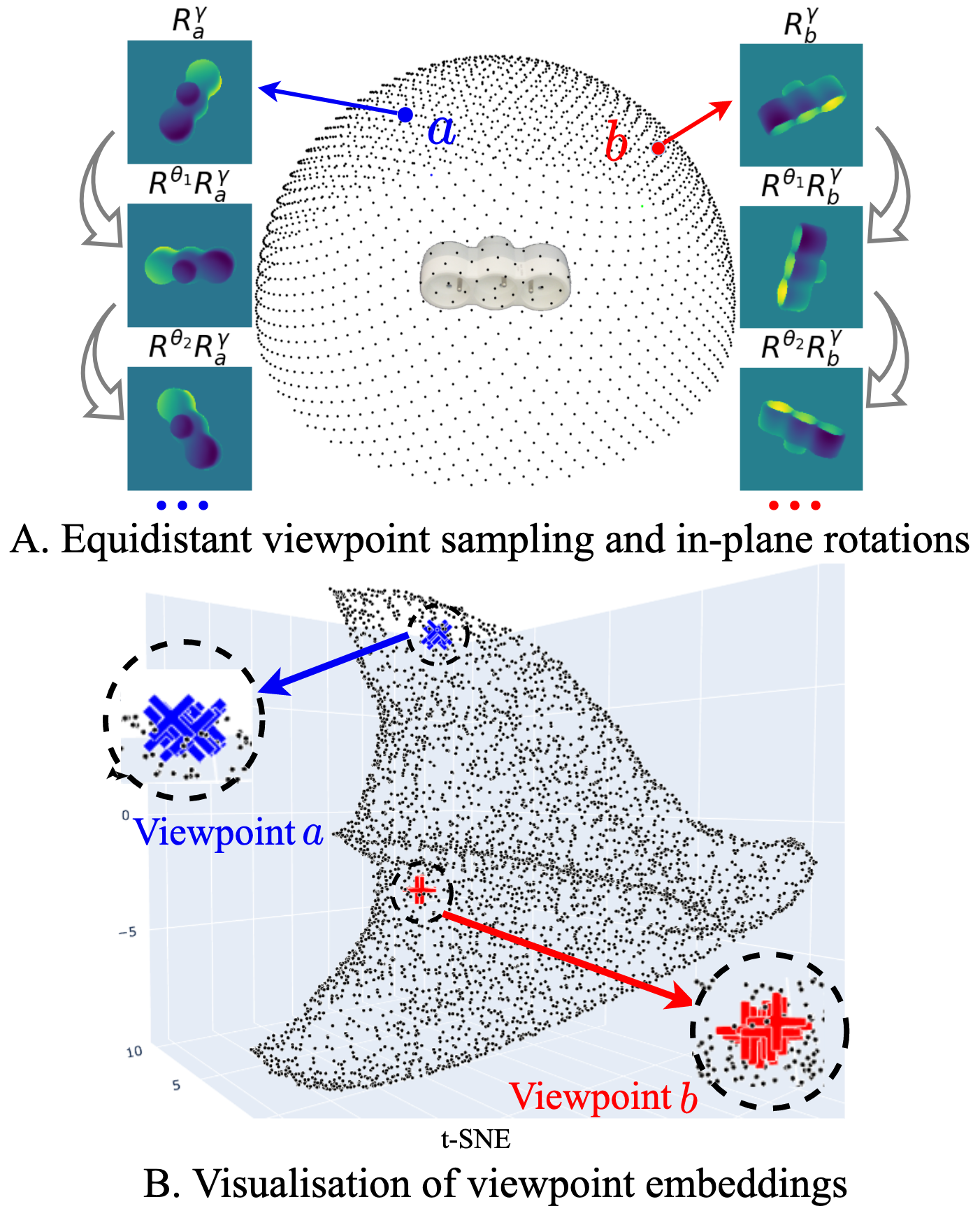}}
\caption{
% Illustrations of the viewpoint sampling and embeddings.
\textbf{A)} 4,000 viewpoints uniformly sampled from a sphere centered on the object (only the upper hemisphere is shown). The in-plane rotations $R^{\theta_i}$ around the camera optical axis are illustrated by synthesizing three examples at viewpoints \textcolor{blue}{$a$} ($R^{\gamma}_a$) and \textcolor{red}{$b$} ($R^{\gamma}_b$). \textbf{B)} The illustration of the proposed viewpoint embeddings using t-SNE \cite{van2008visualizing}, where the blue ('\textcolor{blue}{x}') and red ('\textcolor{red}{+}') points correspond to the embeddings from 10 in-plane rotated views at the viewpoints \textcolor{blue}{$a$} and \textcolor{red}{$b$}, respectively, and the black points represent the remaining viewpoints. 
It can be observed that the embeddings are relatively invariant to the in-plane rotations while varying with respect to the camera viewpoint.
}
\label{fig:sampling}
\end{figure}

The 6D pose of an object refers to a geometric mapping from the object coordinate system to the camera reference frame \cite{hinterstoisser2011multimodal,hodan2018bop}. Most commonly, this transformation is defined in terms of 3D rotation (object orientation) and 3D translation (object location). The ability to infer the object pose is an essential feature for many applications interacting with the environment. For instance, in robotic manipulation \cite{collet2011moped} and augmented reality \cite{marchand2015pose}, the pose is needed for grasping or realistically rendering artificial objects.

In recent works, the object pose estimation problem is commonly approached by either establishing local correspondences between the object 3D model and the observed data \cite{peng2020pvnet,he2021ffb6d,He_2020_CVPR}, or via direct regression \cite{chen2020g2l,shi2021stablepose}. In both cases, the inference models are often optimized and stored separately for each object instance. Such approach quickly turns intractable as the number of object instances grows. Meanwhile, some existing works \cite{wang2019densefusion,Zhou_2021_ICCV} consider building a single model for multiple objects. However, to retain the performance, the model requires expensive re-training every time a new object instance is added to the database. In addition, most of the best-performing methods need annotated real-world training data, which is laborious to obtain. Although some works \cite{sundermeyer2020augmented,sundermeyer2020multi,kehl2017ssd} consider using synthetic examples in training, they suffer from noticeable performance degradation due to the domain gap.

An alternative approach, called LatentFusion, was proposed in \cite{park2019latentfusion}. In this work, they first reconstruct a latent 3D object model from a small set of reference views, and later use the model to infer the 6D pose of the corresponding object from the input image. The main advantage is the ability to add new objects by simply generating new latent models while keeping all network parameters fixed. However, the method is computationally expensive as it is based on iterative optimization at the inference time. Furthermore, LatentFusion is very sensitive to occlusions in the input data, resulting in a significant drop in performance. 

In this paper, we present a new approach, called OVE6D, for estimating the 6D object pose from a single depth image and the object segmentation mask. We further assume to have access to the 3D mesh model of the target object. Similar to LatentFusion, our approach generalizes to new objects without any re-training of model parameters. Moreover, unlike LatentFusion, the proposed method is computationally efficient and robust to occlusions in the input data. In fact, OVE6D obtains the new state-of-the-art results on the challenging T-LESS dataset\cite{hodan2017tless}, surpassing even approaches optimized particularly for this dataset. 

The proposed approach consists of three stages as illustrated in Figure \ref{fig:intro}. First (\cref{fig:intro} A), we train the model parameters using a large number of synthetic 3D object models from the ShapeNet \cite{shapenet2015} dataset. This stage is performed only \emph{once} and the resulting parameters remain fixed in later stages. Second (\cref{fig:intro} B), we convert the 3D mesh models of the target objects into viewpoint codebooks. The conversion is performed \emph{once for each object} and it takes roughly 30 seconds per instance. Finally (\cref{fig:intro} C), the 6D pose is inferred from the input depth image and object segmentation mask. The complete OVE6D model contains less than 4M parameters and requires roughly 50ms to infer the pose for a single object. New, previously unseen, object can be added by simply encoding the corresponding 3D mesh model as in the second stage. 

% repeating the second stage to the new 3D mesh models.

The core of OVE6D is a depth-based object viewpoint encoder that captures the object viewpoint into a feature vector. The encoded representations are trained to be invariant to the in-plane rotation around the camera optical axis, but to be sensitive to the camera viewpoint, as illustrated in Figure \ref{fig:sampling}. At the inference time, we first utilize the viewpoint encodings to determine the camera viewpoint, and subsequently estimate the remaining pose components (camera in-plane rotation and object 3D position) conditioned on the obtained viewpoint. The cascaded pipeline allows compact architectures for each sub-task and enables efficient training using thousands of synthetic objects. %, and allows the model to generalise to new unseen real-world objects. 

To summarize, our key contributions are: 
1) We propose a cascaded object pose estimation framework, which generalizes to previously unseen objects without additional parameter optimization. %does not require any object specific parameter optimisation.
% which does not require any objects specific parameter optimisation. 
%We propose a cascaded framework for inferring the 6D object pose for untrained objects.
% 2) Given a novel object mesh model, it only takes around 30 seconds to extend our framework to perform 6D pose estimation for the object with known object masks.
 2) We propose a viewpoint encoder that robustly captures object viewpoint while being insensitive to the in-plane rotations around the camera optical axis.
 3) We demonstrate the new state-of-the-art results on T-LESS\cite{hodan2017tless}, without using any images from the dataset to train our model.

\section{Related Work}
\paragraph{Pose estimation from RGB data} Most RGB-based object 6D pose estimation methods \cite{brachmann2014learning,pavlakos20176,peng2020pvnet,rad2017bb8,Park_2019_ICCV,tekin2018real,zakharov2019dpod,hodan2020epos} attempt to establish sparse or dense 2D-3D correspondences between the 2D coordinates in the RGB image and the 3D coordinates on the object 3D model surface. The 6D pose is computed by solving the Perspective-n-Point (PnP) problem \cite{lepetit2009epnp}. These methods achieve impressive performance for objects with rich textures providing sufficient local features to determine reliable 2D-3D correspondences. Another intuitive way to estimate the 6D pose is to directly predict the pose parameters using regression or classification, such as \cite{bukschat2020efficientpose,kehl2017ssd,xiang2018posecnn,trabelsi2021pose}. Most of these methods are based on supervised learning and rely on real-world training data with pose annotations. However, recent self-supervised approaches \cite{Sundermeyer_2018_ECCV,sundermeyer2020multi,wang2020self6d,sock2020introducing} take full advantage of the costless synthetic data for training and perform 6D object pose estimation in real-world images at testing time. Similarly, we also adopt self-supervised learning in our work and purely train our network on synthetic data.

\paragraph{Pose estimation from depth data} Some deep learning-based 6D object pose estimation methods use depth-only data. Gao \etal \cite{gao20206d} proposed CloudPose, which is known as the first deep learning system that performs 6D pose regression from the point cloud segments created from the object depth image. Later, Gao \etal proposed CloudAAE \cite{gao2021cloudaae} to improve the generalization of the network trained on synthetic depth data by adopting an augmented autoencoder (AAE)\cite{sundermeyer2020augmented} point cloud based architecture. They argue that the domain gap between the synthetic and the real images is considerably smaller and easier to fill for depth information. Bui \etal\cite{bui2018regression} proposed a multi-task framework combining manifold learning and 3D orientation regression directly from depth images to learn view descriptors. It was further leveraged to either retrieve or regress the 3D pose. Bui \etal\cite{bui2018regression} is most similar to our work, but in this work, we decouple the complete 3D orientation into the viewpoint (out-of-plane rotation) for retrieval and the 2D in-plane rotation for regression. The recent method StablePose \cite{shi2021stablepose} adopts the geometric stability analysis of object patches and directly predicts the patch poses in a stable group to further determine the pose of the object, which achieves state-of-the-art performance on the T-LESS dataset.

\paragraph{Pose estimation from RGB-D data} When both RGB images and depth images are available, the most straightforward utilization of the two modalities is to first perform the initial pose estimation based on RGB images and then further refine with depth images, such as via ICP refinement \cite{hinterstoisser2012model,xiang2018posecnn,Sundermeyer_2018_ECCV,sundermeyer2020multi}. Alternatively, the 2D-3D feature fusion-based approaches \cite{xu2018pointfusion,wang2019densefusion,He_2020_CVPR,he2021ffb6d,chen2020g2l} directly fuse the deep appearance features and the deep geometry features extracted from RGB and depth data by deep neural networks. These methods take full advantage of multi-modal inputs and have achieved high performance on benchmark datasets. Kehl \etal\cite{kehl2016deep} adopt RGB-D patch descriptors extracted by CNN for 6D pose vote casting, which ignores the holistic structures of objects and easily suffers from the poor local textures. In contrast, template-based methods\cite{wohlhart2015learning,zakharov20173d} employ triplet loss to learn view descriptors from entire RGB-D images for object recognition and 3D pose estimation via nearest neighbor search. 

\paragraph{Pose estimation for untrained objects} Many category-level 6D object pose estimation methods \cite{chen2020learning,chen2021fs,chen2020category,tian2020shape,wang2019normalized} have appeared recently and shown good generalization to untrained objects within the same category by assuming that the same canonical pose and similar shape are shared for all instances within a category. In contrast, this assumption is not required by the instance-level object-agnostic methods LatentFusion\cite{park2019latentfusion} and MP-Encoder\cite{sundermeyer2020multi}. The network in LatentFusion \cite{park2019latentfusion} is completely trained on the synthetic data rendered from ShapeNet\cite{shapenet2015}, and it is capable of generalizing to different real-world datasets when a few reference RGB images with pose annotations are provided at testing time. We follow LatentFusion\cite{park2019latentfusion} to train our network on the ShapeNet. However, OVE6D does not require any pose annotation at the training or testing time. In addition, DeepIM\cite{li2018deepim} can perform pose refinement for the untrained objects when the initial pose is given.

\section{Method}

\begin{figure*}[t]
\centering
\includegraphics[width=0.9\linewidth]{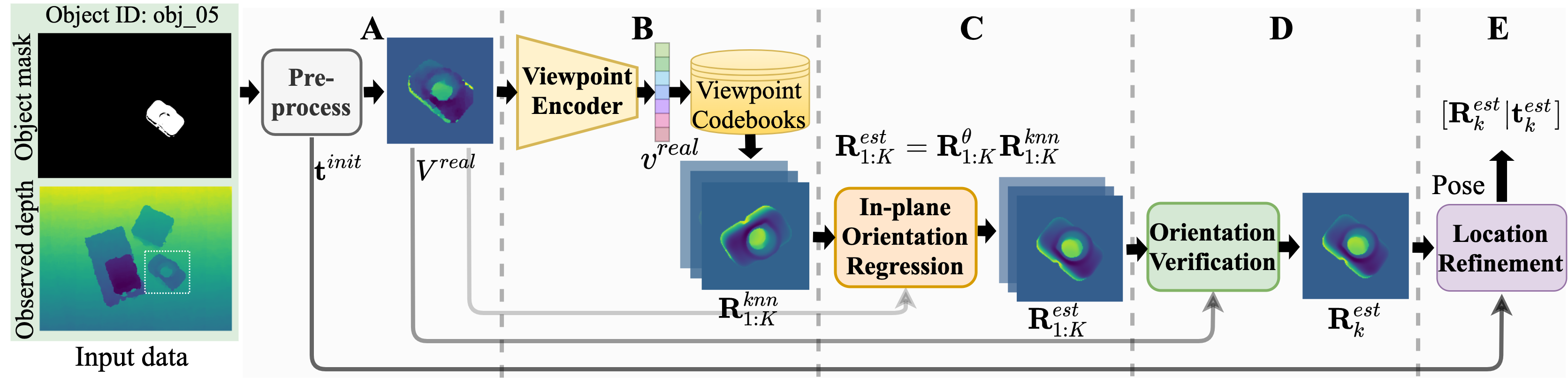}
\caption{
\textbf{The inference pipeline of OVE6D}. The entire system operates in a cascaded manner. First, the raw depth image is pre-processed to $128\times128$ input (\textbf{A}). Second, the object orientation is obtained by performing the viewpoint retrieval (\textbf{B}), in-plane orientation regression (\textbf{C}), and orientation verification (\textbf{D}). Finally, the object location is refined (\textbf{E}) using the obtained orientation and the initial location (\textbf{A}).
}
\label{fig:inference}
\end{figure*}

\begin{figure}[t]
\centering
\includegraphics[width=0.88\linewidth]{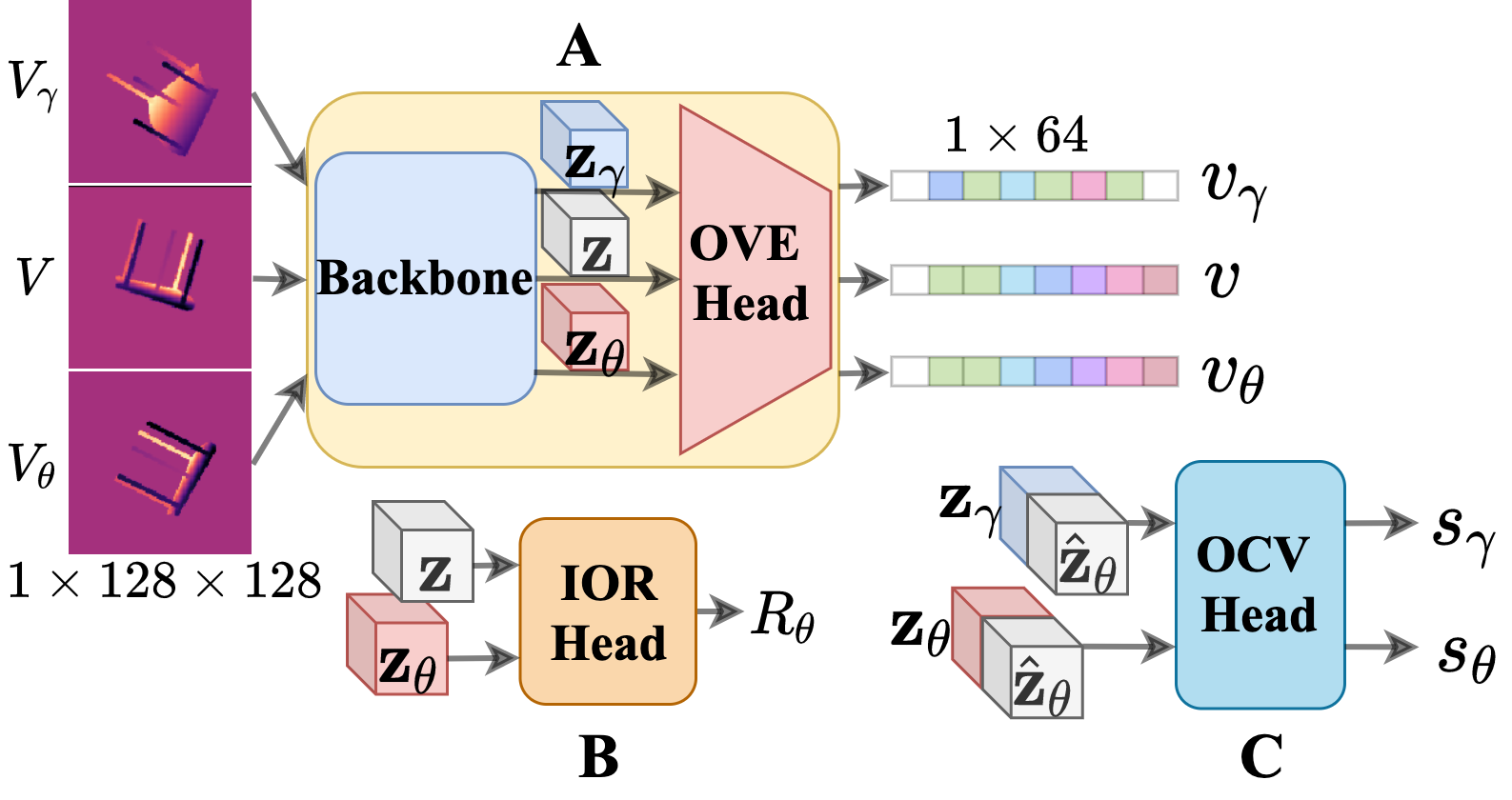}
\caption{
\textbf{Training the networks}. The proposed model contains three sub-networks to be trained. 
The feature maps ($\mathbf{z}, \mathbf{z}_{\theta}, \mathbf{z}_{\gamma}$) are first  extracted from the rendered depth images by the shared backbone network, and then consumed by the Object Viewpoint Encoder (\textbf{OVE}) head (\textbf{A}), the In-plane Orientation Regression (\textbf{IOR}) head (\textbf{B}), and the Orientation Consistency Verification (\textbf{OCV}) head (\textbf{C}). $\hat{\mathbf{z}}_{\theta}$ is transformed from $\mathbf{z}$ with the rotation $R_{\theta}$.}
\label{fig:training}
\end{figure}

In this section, we present our framework called, OVE6D, for 6D object pose estimation. Here, we assume that object IDs are known, 3D mesh models are available and object segmentation masks are provided. 
The task is to predict a rigid transformation from the object coordinate system to the camera coordinate system. Such transformation can be represented with a rotation $R \in SO(3)$ and a translation $t \in R^3$. 
The rotation $R$ can be further factorized into the out-of-plane rotation (viewpoint) $R_{\gamma}$ and the in-plane orientation (rotation around the camera optical axis) $R_\theta$, \ie, $R=R_\theta R_\gamma$, (see \cref{fig:sampling}A).
More details are provided in the supplementary material.

\subsection{Overview}
The OVE6D framework is illustrated in figures \ref{fig:intro}, \ref{fig:inference} and \ref{fig:training}. In the training phase, the model parameters are optimized using the synthetic 3D objects from ShapeNet\cite{shapenet2015}. Next, the object viewpoint codebooks are constructed with the viewpoint encoder module (see \cref{fig:codebook}). At the inference time, we perform the following subtasks in a cascaded fashion. First, an initial location estimate is computed using the input depth image and the object segmentation mask, and applied to preprocess the depth image (see \cref{fig:inference}A). 
Second, we retrieve multiple viewpoint candidates from the object viewpoint codebook (see \cref{fig:inference}B). 
Third, we regress the in-plane 2D rotation with respect to each retrieved viewpoint candidate and obtain a set of complete 3D orientation estimates (see \cref{fig:inference}C). 
Next, we calculate a consistency score for each orientation hypothesis and output one (or more) estimate according to the score values (see \cref{fig:inference}D). Finally, the initial location estimate is refined based on the obtained 3D orientation (see \cref{fig:inference}E). The following subsections outline further details of the model components and the training procedures.

\subsection{Preprocessing}
\label{sec:prep}
First, we calculate and subtract the median distance $d_c$ from the segmented input depth image $D_M$ (obtained by element-wise multiplication of the depth image and the segmentation mask $M$). Next, we calculate the center coordinate $(c_x, c_y)$ of the bounding box enclosing the input segmentation mask, and form an initial estimate of the object 3D location as $t^{init} = K^{-1}[c_x, c_y, d_c]^T$, where $K$ is the camera intrinsic matrix. Finally, we follow LatentFusion\cite{park2019latentfusion} to re-scale and crop $D_M$, according to the estimated location $t^{init}$ to produce $128 \times 128$ pre-processed input depth image for the later stages.

\subsection{Object Viewpoint Encoder}
\label{sec:vp}
The viewpoint encoder is a lightweight neural network comprising of a CNN-based backbone (eight Conv2D + BN layers) and an encoder head $F_{OVE}$ (a single Conv2D, Pooling, and FC layer). The encoder takes the preprocessed $128\times 128$ depth image as input and outputs a feature vector with 64 elements. The feature vector is intended to encode the camera viewpoint, but to be invariant to the in-plane rotation around the camera optical axis.

We train the viewpoint encoder using depth images rendered from ShapeNet \cite{shapenet2015}. The generated samples are organized into triplets $\{V, V_{\theta}, V_{\gamma}\}$, where $V$ and $V_{\theta}$ differ only in terms of in-plane rotation (by angle $\theta$), and $V_{\gamma}$ is rendered from a different camera viewpoint (by angle $\gamma$). The depth images are further embedded into feature representations $\{v, v_{\theta}, v_{\gamma}\}$ using the viewpoint encoder network (see \cref{fig:training}A). The encoder parameters are optimized to rank the representation pairs according to the cosine similarity, \ie, $S(v, v_{\theta}) > S(v, v_{\gamma})$, where $S$ is the cosine similarity function. Thus, the equivalent loss function can be written as,
\begin{equation} \label{eq:vp_loss}
\begin{split}
\ell^{vp} &= \max(S(v, v_\gamma) - S(v, v_\theta)+ m^{vp}_{\lambda}, 0), \\
\end{split}
\end{equation}
where $m^{vp}_{\lambda} \in (0, 1)$ is the ranking margin. %After being purely trained with more than 19,000 synthetic objects from ShapeNet\cite{shapenet2015}, the trained encoder is adopted for real-world data without any fine-tuning. % This is explained in the "training data" part, so might not need to be iterated here. 

\begin{figure*}[t]
\centering
\includegraphics[width=0.80\linewidth]{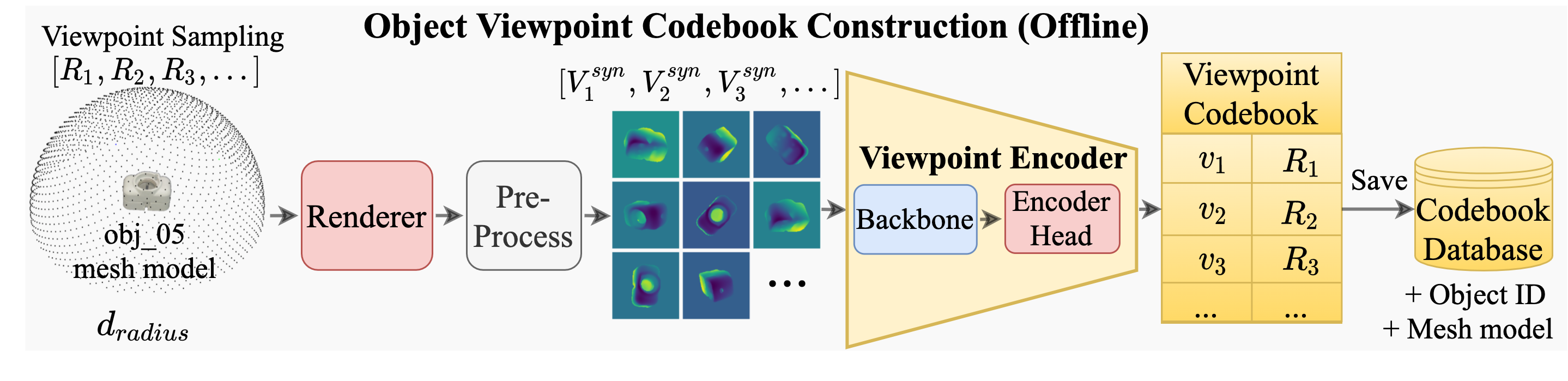}
\caption{
\textbf{Viewpoint codebook construction.} The viewpoints are sampled from a sphere centered on the object mesh model with a radius proportional to the object diameter. 
The viewpoint representations are extracted from the rendered depth images by the viewpoint encoder. 
}
\label{fig:codebook}
\end{figure*}

The trained viewpoint encoder is later used to construct viewpoint codebooks for novel real-world objects. To do this, we first uniformly sample $N$ viewpoints $\{R_i\}^N_{i=1}$ from a sphere centered on the object with the radius of $d_{radius}=f_{base} * d_{diameter}$, where $d_{diameter}$ is the object diameter (obtained from the 3D mesh model) and $f_{base}$ is a distance factor ($5$ in this paper). Then the synthetic noise-free depth images $\{V^{syn}_i\}^N_{i=1}$ are rendered using the sampled viewpoints and the object 3D mesh model. Last, viewpoint representations $\{v_i\}^N_{i=1}$ are extracted from these images (preprocessed as described in \cref{sec:prep}) using the viewpoint encoder and stored into the codebook database along with the object mesh model, as illustrated in Figure \ref{fig:codebook}. The generated viewpoint codebook is a set $\{\{v_i, R_i\}^N_{i=1}, O_{mesh}, O_{id}\}$ that contains the corresponding viewpoint embeddings, rotation matrices, mesh model, and object ID. The entire construction requires approximately 30 seconds per object with $N=4000$ viewpoint samples. 

At the inference time, the object viewpoint representation $v^{real}$ is first extracted from the preprocessed depth image $V^{real}$ using the viewpoint encoder. Then, we utilize $v^{real}$ to compute the cosine similarity scores with all entries in the corresponding viewpoint codebook (indexed with the known object ID). The entry $\{v^{knn}, R^{knn}\}$ with the highest similarity between $v^{real}$ and $ v^{knn}$ is selected as the nearest viewpoint for $V^{real}$. Optionally, we can select multiple candidate entries $\{v^{knn}_{k}, R^{knn}_{k}\}^K_{k=1}$ from the codebook to obtain a pool of $K$ viewpoint hypotheses according to the descending cosine similarity scores, as shown in Figure \ref{fig:inference}B. 
% The entire retrieval procedure takes around 6 ms per object.

\subsection{In-plane Orientation Regression}
\label{sec:Rz}
% (in term of orthogonal projection)
%The in-plane rotation around the camera optical axis approximately reduces to estimating a 2D image rotation angle with a known viewpoint. 
Once the viewpoint is known, the in-plane rotation around the camera optical axis can be approximated using a 2D rotation of the depth image (exact for orthographic camera). To this end, we build a regression network by appending a regression head $F_{IOR}$ (one Conv2D and two consecutive FC layers) to the backbone shared with the viewpoint encoder. This module takes a pair of feature maps $\{\mathbf{z}, \mathbf{z}_{\theta}\} \in R^{c\times h \times w}$ of the same viewpoint with varying in-plane orientations (intra-viewpoint) as input and regresses the relative in-plane rotation angle $\theta$ (represented as a matrix $R_{\theta}$), as shown in Figure \ref{fig:training}B. 

We train this module to minimize the discrepancy between the depth images transformed by the ground-truth rotation matrix $\hat{R}_\theta$ and the predicted $R_{\theta}$. Here, we employ a negative logarithmic cosine similarity to measure the discrepancy, written as
\begin{equation} \label{eq:rot_loss}
\begin{split}
S_{cos} &= S(\mathbbm{F}(T_{R_{\theta}}(V)), ~ \mathbbm{F}(T_{\hat{R}_\theta}(V)), \\
\ell^{\theta} &= -\log((1.0 + S_{cos})~/ ~2.0),\\
\end{split}
\end{equation}
where $\mathbbm{F}$ refers to the flattening operation, 
$T_{R_{\theta}}$ represents the 2D spatial transformation
% \footnote{A geometric transformation of the image coordinate system (\eg, re-sampling the image according to the transformed coordinate system).}
\cite{jaderberg2015spatial} with $R_{\theta}$, 
and $V$ is the viewpoint depth image. 
% The module structure is provided in the supplementary material.

At the inference time, we first use the shared backbone network to extract a feature map pair $\{\mathbf{z}^{real}, \mathbf{z}^{knn}_k\}$ from the preprocessed depth image pair $\{V^{real}, V^{knn}_k\}$, where $V^{knn}_k$ is the synthesized depth image using the retrieved viewpoint $R^{knn}_k$. Next, the regression module takes the feature map pair to estimate the relative 2D rotation matrix $R^{\theta}_k = F_{rot}(\mathbf{z}^{real}, \mathbf{z}^{knn}_k)$ to produce the complete 3D orientation estimate via $R^{est}_k=R^{\theta}_kR^{knn}_k$.  In addition, the in-plane orientation regression can be concurrently performed for several retrieved viewpoints to obtain multiple 3D orientation hypotheses $\{R^{est}_{k}\}^K_{k=1}$, as shown in Figure \ref{fig:inference}C.

\subsection{Orientation Consistency Verification}
\label{sec:css}
Multiple complete 3D orientation hypotheses $\{R^{est}_k\}^K_{k=1}$ can be derived from the previous modules, as described in Section \ref{sec:Rz}. To rank the candidates, we adopt an orientation verification module that estimates the consistency between the candidates and the actual object orientation depicted in $V^{real}$. Similar to the regression module, the verification module is built by appending a verification head $F_{OCV}$ (two Conv2D layers, a Pooling and FC layer) to the shared backbone.

At the training time, we adopt a ranking-based loss to optimize this module. As shown in Figure \ref{fig:training}C, the feature map $\mathbf{z}$ is first spatially transformed using the in-plane rotation, \ie $\hat{\mathbf{z}}_{\theta}=T_{R_{\theta}}(\mathbf{z})$ where $T_{R_{\theta}}$ is the spatial transformation\cite{jaderberg2015spatial} with $R_{\theta}$. Then, we separately concatenate $\hat{\mathbf{z}}_{\theta}$ with $\mathbf{z}_{\gamma}$ and $\mathbf{z}_{\theta}$ along the feature channel dimension, \ie, $[\hat{\mathbf{z}}_{\theta}; \mathbf{z}_{\gamma}]$ and $[\hat{\mathbf{z}}_{\theta}; \mathbf{z}_{\theta}]$, where $[;]$ denotes the concatenation, and feed them into $F_{OCV}$ to produce the consistency scores $s_{\gamma}$ and $s_{\theta}$. The equivalent loss function can be written as,
\begin{equation} \label{eq:css_loss}
\begin{split}
\ell^{css} &= \max(s_{\gamma} - s_{\theta} + m^{css}_{\lambda}, 0), \\
\end{split}
\end{equation}
where
% \begin{equation}
% \begin{split}
% s^{css}_{\gamma} &= Sigmoid(f_{css}([\hat{\mathbf{f}}_{\theta}; \mathbf{f}_{\gamma}])), \\
% s^{css}_{\theta} &= Sigmoid(f_{css}([\hat{\mathbf{f}}_{\theta}; \mathbf{f}_{\theta}])), \\
% \end{split}
% \end{equation}
$m^{css}_{\lambda} \in (0, 1)$ is the ranking margin.
% , $\{s^{css}_{\gamma}, s^{css}_{\theta}\}$ is the output of the network $f_{css}$.
% $[;]$ donates the concatenation, and $\{\mathbf{f}, \mathbf{f}_{\gamma}, \mathbf{f}_{\theta}\} \in R^{c\times h\times w}$ are the feature maps extracted by the shared backbone network.

%During the inference, we transform the feature map $\mathbf{z}^{knn}_k$ of the retrieved viewpoint using the estimated in-plane rotation $R^{\theta}_k$ and feed it into the verification head $F_{css}$ with the feature map $\mathbf{z}^{real}$ from the observed depth image, as shown in Figure \ref{fig:inference}D.
%In this way, we obtain a consistency score for each 3D orientation hypothesis. According to the estimated scores, we rank all hypotheses $\{R^{est}_k\}^K_{k=1}$ in descending order and select the top $P \in [1, K]$ orientation proposals $\{R^{est}_p\}^{P}_{p=1}$ as the output for robustness.

During the inference, we transform the feature map $\mathbf{z}^{knn}_k$, from the retrieved viewpoint, using the estimated in-plane rotation $R^{\theta}_k$ and feed it to the verification head $F_{css}$ along with the feature map $\mathbf{z}^{real}$ from the observed depth image, as shown in Figure \ref{fig:inference}D.
In this way, we obtain a consistency score for each 3D orientation hypothesis. According to the estimated scores, we rank all hypotheses $\{R^{est}_k\}^K_{k=1}$ in descending order and select the top $P \in [1, K]$ orientation proposals $\{R^{est}_p\}^{P}_{p=1}$ as the output.

\subsection{Location Refinement}
\label{refine_t}
% The initial location $t^{init}$ of the object is computed from the object segmentation mask $M$ and the depth image $D_M$, which could lead to a position shift along the camera optical axis (z-axis) due to the self-occlusion of the object. To this end, we adopt a refinement step to alleviate this problem. 

We further refine the initial location estimate $t^{init}$ based on the obtained 3D orientation. Specifically, we first synthesize a depth image $D^{est}_{p}$ using the object mesh model and the pose $[R^{est}_p|t^{init}]$, where $R^{est}_p$ is the 3D orientation obtained in Section \ref{sec:css}. Next, we estimate the 3D centroid  $t^{syn}_p$ of the object in the depth image $D^{est}_{p}$, as described in Section \ref{sec:prep}. Furthermore, we calculate an offset $t'_{\Delta} = t^{init} - t^{syn}_p$ which can be regarded as the position offset caused by the self-occlusion of the object in the current orientation $R^{est}_p$. We assume that $t'_{\Delta}$ is approximately equal to $t_{\Delta} = t^{est} - t^{init}$, which allows us to obtain the final 3D location estimate $t^{est}_p = 2t^{init} - t^{syn}_p$ for the object with the $p^{th}$ orientation proposal, as $t^{est}_p - t^{init} = t^{init} - t^{syn}_p$.

% We obtain the 3D translation estimate $t^{est}$ using the following two-stage procedure. In the first stage, we calculate an initial translation $t^{init}$ by estimating the 3D centroid of the object from the input depth image $D^{real}_{test}$ and the object mask, as described in \cite{park2019latentfusion}. In the second stage, we first synthesize a depth image $D^{est}_{p}$ using the object 3D model and the pose $[R^{est}_p|t^{init}]$, where $R^{est}_p$ is the 3D orientation obtained in \cref{sec:css}. Again, we obtain a translation estimate $t^{syn}_p$ from the $D^{est}_{p}$ as in the first stage, and then use it to calculate a translation offset $t'_{\Delta} = t^{init} - t^{syn}_p$. We assume that $t'_{\Delta}$ is roughly equal to $t_{\Delta} = t^{est} - t^{init}$, which allows us to obtain the final 3D translation estimate $t^{est}_p = 2 * t^{init} - t^{syn}_p$ for the $p^{th}$ orientation proposal, as $t^{est}_p - t^{init} = t^{init} - t^{syn}_p$. We provide further details in the supplementary material. 
\subsection{Pose Hypothesis Selection and Refinement}
\label{sec:multi_pose}
As presented in previous sections, we may obtain multiple orientation proposals, each of which results in one pose hypothesis. We calculate the following quality measure for each pose hypothesis,
\begin{equation} 
\label{eq:verify}
q_p =\dfrac{1}{m_p}\sum{ \mathbbm{I} (|D^{syn}_p - D_M| > 0.1d)}
\end{equation}
where $\mathbbm{I}$ represents the indicator function, $D_M$ is the segmented object depth image (obtained in \cref{sec:prep}), $D^{syn}_p$ is the rendered depth image with the pose hypothesis $\{R^{est}_p|t^{est}_p\}$, $m_p$ is the total number of pixels belonging to the object in $D^{syn}_p$, and $d$ is the diameter of the target object and $q_p$ represents the ratio of the outlier pixels. The pose hypothesis with the lowest $q_p$ value among $\{q_p\}^P_{p=1}$ is selected as the final output pose.

Furthermore, the obtained pose can be optionally refined using the Iterative Closest Point (ICP) algorithm ICP \cite{chen1992object,zhang1994iterative}. The ICP refinement can be done before or after the pose hypothesis selection as shown in the experiments. 
% to refine our results, and it can be performed before or after the pose hypothesis selection as shown in the experiments.

\subsection{Combined Loss Function}
The entire network consists of a single shared backbone with three head branches and is trained in an end-to-end fashion.
The overall training loss is 
\begin{equation} \label{eq:total_loss}
\small
\begin{split}
L &= \frac{1}{bs}\sum^{bs}_{i}(\lambda_{1} \ell^{vp}_i + \lambda_{2} \ell^{css}_i + \lambda_{3} \ell^{\theta}_i), \\
\end{split}
\end{equation}
where $bs$ is the batch size, and $\lambda_1, \lambda_2$ and $\lambda_3$ are weighting parameters. In our experiments, we set the ranking margins $ m^{vp}_{\lambda} = m^{css}_{\lambda}=0.1$ and the weights $\lambda_{1}=100, \lambda_{2}=10, \lambda_{3}=1$. 
% Further implementation details are provided in the supplementary material. 

\subsection{Implementation Details}
We implement the method using the PyTorch\cite{paszke2019pytorch} framework and utilize Adam solver\cite{kingma2014adam} with the cosine annealing learning rate starting from $1 \times 10^{-3}$ to $1 \times 10^{-5}$ and weight decay $1 \times 10^{-5}$ for training 50 epochs (around three days) on a single Nvidia RTX3090 GPU.

% \paragraph{Pre-processing} The input depth image is pre-processed before being fed into the orientation estimation network. First, the input depth is multiplied with the object segmentation mask and normalized by subtracting the mean depth value. Second, we follow LatentFusion \cite{park2019latentfusion} to zoom in on this depth image and crop it using a dilated object bounding box with the size of $128 \times 128$. Please refer to \cite{park2019latentfusion} for more details.

\paragraph{Training Data} Our training data is generated from the public 3D shape dataset \cite{shapenet2015}. Following LatentFusion\cite{park2019latentfusion}, we exclude large objects for efficient data loading and obtain $\sim$19k shapes over the original 52,274 shapes. For each object, we first randomly sample 16 anchor viewpoints $\{R_i\}^{16}_{i=1}$ distributed on a sphere centered on the object. Next, we separately apply a random in-plane rotation $R^\theta_i$ ($R^\theta_i R_i$) and a random out-of-plane rotation $R^\gamma_i$ ($R^\gamma_i R_i$) for each anchor viewpoint, which results in a batch of viewpoint triplets for a single object. We randomly select eight objects each time and form a training batch with the size of 128. The Pyrender\cite{pyrender} library is employed to synthesize the corresponding depth images from these sampled viewpoints. Similar to \cite{park2019latentfusion}, we use data augmentation techniques to improve the generalization of the network.
% More details are provided in the supplementary material.

%We use the predicted and the ground truth object segmentation masks in our experiments. We initialize the Mask-RCNN\cite{he2017mask} network with the pre-trained weights from Detectron2\cite{wu2019detectron2} and fine-tune it only with 50,000 synthetic RGB images provided by BOP\cite{hodavn2020bop}. We use the class labels provided by Mask-RCNN as object IDs in the experiments.
% To narrow the domain gap, we first freeze the backbone of the Mask-RCNN initialized with the pre-trained weights (\eg trained on MS-COCO dataset \cite{lin2014microsoft}) and train 50,000 iterations on the synthetic training data. Then, we unfreeze the backbone and train another 50,000 iterations on the same training data. 
More details are provided in the supplementary material.

\begin{table}[t]
\scriptsize
\centering
\renewcommand{\arraystretch}{1.2}
\begin{tabular}{c | c | c c c c}
\hline
{\shortstack{~\\General-\\ization}} & {\shortstack{~\\Train \\ Data}} & {\shortstack{~\\Method\\~}} & {\shortstack{~\\Input\\~}} & {\shortstack{~\\ICP\\~}} & {\shortstack{VSD\\(\%)}} \\ 
\hline\hline
\multirow{7}{*}{\shortstack{Single \\ Trained \\Object}} 
& \multirow{5}{*}{\shortstack{Real\\(+Syn.)}}%{}
& Pix2Pose\cite{Park_2019_ICCV} & RGB &  ~ & 29.5 \\
& ~ & PVNet\cite{peng2020pvnet} & RGB & ~ & 40.4 \\ 
& ~ & PPFNet\cite{deng2018ppfnet} & D & ~ & 49.0 \\
& ~ & PointNet++\cite{qi2017pointnet++} & D & ~ & 54.0 \\
& ~ & StablePose\cite{shi2021stablepose} & D & ~ & \bf{73.0} \\
\cline{2-6}
& \multirow{2}{*}{\shortstack{Syn.\\Only}}
& AAE\cite{sundermeyer2020augmented} & RGB & ~ & 19.3 \\
& ~ & AAE\cite{sundermeyer2020augmented} & RGBD & \cmark & 68.6\\
\hline\hline
\multirow{5}{*}{\shortstack{Multi- \\ Trained \\ Objects}} 
& \multirow{3}{*}{\shortstack{Real\\(+Syn.)}}%{}
& CosyPose\cite{labbe2020cosypose} & RGB & ~ & 63.8 \\
& ~ & DenseFusion\cite{wang2019densefusion} & RGBD & ~ & 10.0 \\
& ~ & Kehl-16\cite{kehl2016deep} & RGBD & \cmark & 24.6\\
\cline{2-6}
& \multirow{2}{*}{\shortstack{Syn.\\Only}}
& MP-Encoder\cite{sundermeyer2020multi} & RGB & ~ & 20.5 \\ 
& ~ & MP-Encoder\cite{sundermeyer2020multi} & RGBD & \cmark & \bf{69.5} \\
\hline\hline
\multirow{8}{*}{\shortstack{ Universal \\Objects}} 
& \multirow{2}{*}{---}
& DrostPPF\cite{drost2010model} & D & ~ & 57.0 \\
& ~ & VidalPPF\cite{Vidal2018AMF} & D & ~ & 72.0 \\
\cline{2-6}
& \multirow{7}{*}{\shortstack{Syn.\\Only}}
& LatentFusion \cite{park2019latentfusion} & RGBD & ~  & -- \\
& ~ & OVE6D({GT}) & D & ~ & \textit{85.1}\\%84.5 \\ %81.54\\
& ~ & OVE6D({GT}) & D & \cmark & \textit{89.0}\\%86.4 \\ %83.85\\
& ~ & OVE6D({GT}){\ddag} & D & \cmark & \bf{\textit{91.0}}\\%\bf{89.7}\\ % 87.14\\
& ~ & OVE6D(MRCNN) & D & ~ & 69.4\\% 69.2 \\ %68.3 \\ %69.30\\
& ~ & OVE6D(MRCNN) & D & \cmark & 73.1\\%72.4 \\ %71.4\\% 71.56\\
& ~ & OVE6D(MRCNN){\ddag} & D & \cmark & \textbf{74.8}\\%\bf{75.0} \\ %\bf{74.2}\\
\end{tabular}
\caption{Evaluation on \textbf{T-LESS}. We report the average {VSD} recall. 
{\ddag} represents the ICP refinement performed for all pose proposals before selection. 
We highlight the best performance in bold for each group.
MRCNN and {GT} indicate using the masks provided by Mask-RCNN and the ground truth, respectively. 
}
\label{table:tless}
\end{table}

\section{Experiments}
\paragraph{Datasets}
OVE6D is evaluated on three public benchmark datasets: LINEMOD\cite{hinterstoisser2011multimodal}, Occluded LINEMOD\cite{brachmann2014learning}, and T-LESS\cite{hodan2017tless}. LINEMOD (LM) is one of the most popular datasets for single object 6D pose estimation, and it contains RGB-D images and 3D object models of 13 texture-less household objects in cluttered scenes. We construct the test set following the previous works \cite{wang2019densefusion,tekin2018real}. We note that the training set of LINEMOD is completely ignored as OVE6D is fully trained using ShapeNet.  
%However, the training set is completely ignored in our experiments as OVE6D is purely trained on ShapeNet. 
% The final test set contains $\sim$ 13.4k images. 
Occluded LINEMOD (LMO) is a subset of LINEMOD for multi-object 6D pose estimation and contains eight annotated objects in 1214 testing images with heavy occlusions. T-LESS is a challenging dataset including 30 texture-less and symmetric industrial objects with highly similar shapes. The evaluation is performed on the PrimeSense test set, and we report the results for a single object per class following the protocol specified in the BOP challenge\cite{hodan2018bop}.

\begin{table}[t]
\scriptsize
\centering
\renewcommand{\arraystretch}{1.1}
\begin{tabular}{c | c | c c c c c}
\hline
{\shortstack{~\\General-\\ization}} & {\shortstack{~\\Train\\Data}} & {\shortstack{~\\Method\\~}} & {\shortstack{~\\Input\\~}} & {\shortstack{~\\ICP\\~}}  & {\shortstack{ADD \\ (-S)(\%)}} \\ 
\hline\hline
\multirow{7}{*}{\shortstack{Single \\Trained \\Object}} 
& \multirow{4}{*}{\shortstack{Real\\(+Syn.)}}
& Self6D \cite{wang2020self6d} & RGBD & ~ & 86.9 \\% with real data but without real label
& ~ & G2LNet\cite{chen2020g2l} & RGBD & ~ & 98.7 \\
& ~ & PVN3D\cite{He_2020_CVPR} & RGBD & ~ & 99.4 \\
& ~ & FFB6D\cite{he2021ffb6d} & RGBD & ~ & \bf{99.7} \\
\cline{2-6}
& \multirow{3}{*}{\shortstack{Syn.\\Only}}
& Self6D\cite{wang2020self6d} & RGBD & ~ & 40.1 \\
& ~ & AAE\cite{sundermeyer2020augmented} & RGBD & \cmark & 71.6 \\
& ~ & SSD6D\cite{kehl2017ssd} & RGBD & \cmark & 90.9 \\
\hline\hline
\multirow{7}{*}{\shortstack{Multi- \\ Trained \\ Objects}} 
& \multirow{4}{*}{\shortstack{Real\\(+Syn.)}}%{}
& DenseFusion\cite{wang2019densefusion} & RGBD & ~ & 94.3 \\
& ~ & PR-GCN\cite{Zhou_2021_ICCV} & RGBD & ~ & \bf{99.6} \\
& ~ & CloudAAE({GT})\cite{gao2021cloudaae} & D & ~ & \textit{86.8 }\\
& ~ & CloudAAE({GT})\cite{gao2021cloudaae} & D & \cmark & \textit{95.5}\\
\cline{2-6}
& \multirow{3}{*}{\shortstack{Syn.\\Only}} 
& CloudPose ({GT})\cite{gao20206d} & D & \cmark & \textit{75.2} \\
& ~ & CloudAAE({GT})\cite{gao2021cloudaae} & D & ~ & \textit{82.1} \\
& ~ & CloudAAE({GT})\cite{gao2021cloudaae} & D & \cmark & \textit{92.5} \\
\hline\hline

\multirow{7}{*}{\shortstack{ Universal \\Objects}} 
% & \multirow{2}{*}{---}
% & DrostPPF\cite{drost2010model} & D & ~ & 82.0 & -  \\
% & ~ & VidalPPF\cite{Vidal2018AMF} & D & ~ & \bf{90.7} & -  \\
% \cline{2-7}
& \multirow{7}{*}{\shortstack{Syn.\\Only}}
& LatentFusion({GT})\cite{park2019latentfusion} & RGBD & ~ & \textit{87.1} \\
& ~ & OVE6D({GT}) & D & ~ & \textit{96.4}\\%92.2 \\
& ~ & OVE6D({GT}) & D & \cmark & \textit{98.3}\\% 95.7 \\
& ~ & OVE6D({GT}){\ddag} & D & \cmark & \bf{\textit{98.7}}\\%\bf{97.6} \\
& ~ & OVE6D(MRCNN) & D & ~ & 86.1\\%83.6\\%84.1\\ %78.3\\
% & ~ & OVE6D(MRCNN){\ddag} & D & \cmark & 89.8 & ~ \\%90.4\\%90.8\\ %88.2\\
& ~ & OVE6D(MRCNN) & D & \cmark & 91.4\\%89.0\\%88.2 \\ %84.6\\
& ~ & OVE6D(MRCNN){\ddag} & D & \cmark & \bf{92.4}\\%90.4\\%90.8\\ %88.2\\
\end{tabular}
\caption{Evaluation on \textbf{LINEMOD}. We report the average {ADD(-S)} recall. 
We highlight the best performance in bold for each group.
{\ddag} represents ICP refinement performed for all pose proposals before selection. 
MRCNN and {GT} indicate using the masks provided by Mask-RCNN and the ground truth, respectively.   
% \cite{gao20206d} is cited from \cite{gao2021cloudaae}.
}
\label{table:lm}
\end{table}

\paragraph{Segmentation Mask} The object segmentation mask is one of the inputs to the proposed pose estimation method. In the experiments, we obtain the masks using off-the-shelf implementation of Mask-RCNN \cite{he2017mask} from the Detectron2 \cite{wu2019detectron2} library. We train Mask-RCNN using a large set of synthetic images generated from the object models. We use the class labels provided by Mask-RCNN as object IDs in the experiments. In addition, we also report the results for ground truth segmentation masks. 

\paragraph{Metrics and Configurations} 
We follow prior works\cite{Sundermeyer_2018_ECCV,sundermeyer2020multi,hinterstoisser2012model} and report the results in terms of two standard 6D pose estimation metrics ADD(-S)\cite{hinterstoisser2012model, hodan2018bop} (for LM and LMO) and VSD\cite{hodan2018bop} (for T-LESS). 
% We use $e_{ADD(-S)}<0.1d$ where $d$ is the object diameter and the standard VSD recall metric at $e_{vsd}<0.3$ with tolerance $\tau =20mm$, $\delta=15mm$ and object visibility ${>10\%}$. 
Please refer to \cite{hinterstoisser2012model,hodan2018bop} for more details. Furthermore,
we use $N=4000$, $K=50$, and $P=5$ for OVE6D if not otherwise stated. 
%We report the results with and without the point-to-plane ICP \cite{chen1992object,zhang1994iterative} refinement.
% \paragraph{Configurations}

% For the T-LESS dataset, we directly adopt the predicted object segmentation masks from MP-Encoder\cite{sundermeyer2020multi} (obtained with Mask-RCNN). For LM and LMO, the segmentation masks are obtained using Mask-RCNN \cite{he2017mask} and only the viewpoints from the upper hemisphere are used.
% We accept all segmentation results above a chosen threshold and verify these predictions based on viewpoint embeddings cosine similarities. 

\subsection{Comparison with the state-of-the-art}
% We focus on pose estimation part here and assume that object segmentation masks are provided (if required). 
% The comparison is categorized into three groups in terms of generalization. Specifically, \textbf{single model per object} (object-specific): a single model is explicitly trained for each object; \textbf{single model per dataset} (dataset-specific): a single model is trained for each dataset; \textbf{single universal model}: a single trained model for all objects across multiple datasets.

We compare OVE6D against the recent (mainly learning-based) pose estimation works using the popular T-LESS, LINEMOD, and Occluded LINEMOD datasets. We categorize the methods into three main groups in terms of generalization. The methods in the first and second groups train a separate model for each individual object or a model for multiple objects, respectively. The third group consists of methods that do not require any dataset-specific training, other than obtaining the 3D models of the target objects. We further split these groups into approaches that use real and/or synthetic data during model training. Notably, OVE6D and LatentFusion belong to the third category and use only synthetic data for training.

%The compared methods are categorized into three groups in terms of generalization. Specifically,
%\textbf{single trained object}: a single model is exclusively trained for a single object; \textbf{ multiple trained objects}: a single model is trained for multiple objects in the dataset; \textbf{universal objects}: a single model is trained for all objects across multiple datasets. Our model belongs to the last one.

% and the training data used for object 6D pose estimation, \ie synthetic data only or requiring real data for training. 
% We 

\paragraph{T-LESS} The results for OVE6D and the baseline methods are reported in Table \ref{table:tless} in terms of VSD metric. We do not report the results for LatentFusion as it does not perform well due to the occlusions. Note that all other learning-based methods are trained on the T-LESS dataset, unlike OVE6D, which is trained on ShapeNet. Nevertheless, OVE6D still achieves the state-of-the-art performance. In particular, OVE6D with ICP improves over the recent state-of-the-art method StablePose \cite{shi2021stablepose} by noticeable margin of $1.8$\%, regardless that StablePose trains a separate model for each object using real-world examples with pose annotations. The non-learning based VidalPPF\cite{Vidal2018AMF} also performs well, but the approach is computationally expensive. Moreover, OVE6D achieves 91\% recall when using the ground truth masks, indicating the potential for performance improvement with better segmentation masks. The results indicate that OVE6D is particularly suitable for texture-less and symmetric industrial objects.

\begin{table}[t]
\begin{center}
\scriptsize
\renewcommand{\arraystretch}{1.1}
\begin{tabular}{c | c | c c c c c}
\hline
{\shortstack{~\\General-\\ization}} & {\shortstack{~\\Train\\Data}} & {\shortstack{~\\Method\\~}} & {\shortstack{~\\Input\\~}} & {\shortstack{~\\ICP\\~}} & {\shortstack{ADD \\ (-S)(\%)}} \\ 
\hline\hline
\multirow{5}{*}{\shortstack{Single \\ Trained \\Object}} 
& \multirow{5}{*}{\shortstack{Real\\(+Syn.)}}%{}
& PVNet\cite{peng2020pvnet} & RGB & ~ & 42.4 \\
& ~ & PVN3D\cite{He_2020_CVPR} & RGBD & ~ & 63.2 \\
& ~ & FFB6D\cite{he2021ffb6d} & RGBD & ~ & 66.2 \\
& ~ & PVNet\cite{peng2020pvnet} & RGBD & \cmark & \bf{79.0} \\
% & ~ & StablePose\cite{shi2021stablepose} & D & ~ & 63.0{*}\\
\hline\hline
\multirow{9}{*}{\shortstack{Multi- \\Trained \\ Objects}} 
& \multirow{6}{*}{\shortstack{Real\\+Syn.}}%{}
& PoseCNN\cite{xiang2018posecnn} & RGB & ~ & 24.9 \\
% & ~ & CosyPose\cite{labbe2020cosypose} & RGB & ~ & 68.0{*} \\
& ~ & PR-GCN\cite{Zhou_2021_ICCV} & RGBD & ~ & 65.0 \\
& ~ & PoseCNN\cite{xiang2018posecnn} & RGBD & \cmark & \bf{78.0}  \\
& ~ & CloudAAE({GT})\cite{gao2021cloudaae} & D & ~ & \textit{58.9} \\
& ~ & CloudAAE({GT})\cite{gao2021cloudaae} & D & \cmark & \textit{66.1} \\
\cline{2-6}
& \multirow{3}{*}{\shortstack{Syn.\\Only}}
& CloudPose ({GT})\cite{gao20206d} & D & \cmark  & \textit{44.2} \\
& ~ & CloudAAE({GT})\cite{gao2021cloudaae} & D & ~ & \textit{57.1} \\
& ~ & CloudAAE({GT})\cite{gao2021cloudaae} & D & \cmark & \textit{63.2} \\
\hline\hline
\multirow{8}{*}{\shortstack{Universal \\ Objects}} 
% & \multirow{2}{*}{---}
% & DrostPPF\cite{drost2010model} & D & ~ & 55.4 & - \\
% & ~ & VidalPPF\cite{Vidal2018AMF} & D & ~ & 61.9 & - \\
% \cline{2-6}
& \multirow{8}{*}{\shortstack{Syn.\\Only}}
& LatentFusion\cite{park2019latentfusion} & RGBD & ~  & - \\
& ~ & OVE6D({GT}) & D & ~ & \textit{70.9}\\%60.3\\
& ~ & OVE6D({GT}) & D & \cmark & \textit{80.0}\\%69.0\\
& ~ & OVE6D({GT}){\ddag} & D & \cmark & \bf{\textit{82.5}}\\%\bf{74.8}\\
& ~ & OVE6D(MRCNN) & D & ~  & 56.1\\%54.5 \\ %53.6 \\ %46.3\\
% & ~ & OVE6D(MRCNN) & D & ~ & 80.5{*} \\
& ~ & OVE6D(MRCNN) & D & \cmark  & 70.3 \\%67.6\\%66.3\\%58.8\\
& ~ & OVE6D(MRCNN){\ddag} & D & \cmark & \bf{72.8} \\%70.5\\%71.0 \\ %\bf{65.3}\\
% & ~ & OVE6D(MRCNN){\ddag} & D & \cmark & ~ & \bf{72.7}\\%70.5\\%71.0 \\ %\bf{65.3}\\
\end{tabular}
\caption{Evaluation on \textbf{Occluded LINEMOD}. We report the average {ADD(-S)} and VSD recalls.
{\ddag} depicts ICP refinement performed for all pose proposals before selection. 
We highlight the best performance in bold for each group.
MRCNN and {GT} indicate using the masks provided by Mask-RCNN and the ground truth, respectively.
}
\label{table:lmo}
\end{center}
\end{table}

\begin{figure*}[t]
\centering
  \begin{subfigure}{0.33\linewidth}
    % {\includegraphics[width=1.\textwidth]{vp_dve_hist_rand_R50.png}}
    {\includegraphics[width=1.\textwidth]{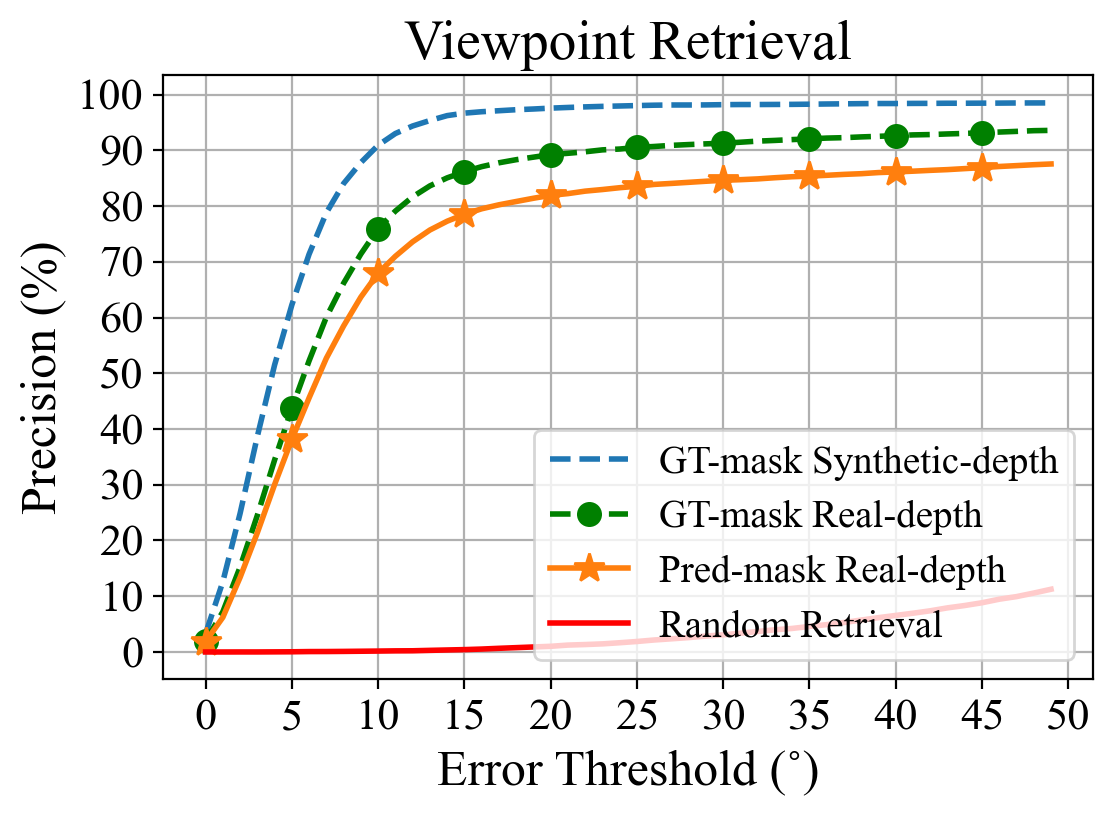}}
  \end{subfigure}
  \begin{subfigure}{0.33\linewidth}
%   {\includegraphics[width=1.\textwidth]{Rz_dve_icp20_R50_2.png}}
{\includegraphics[width=1.\textwidth]{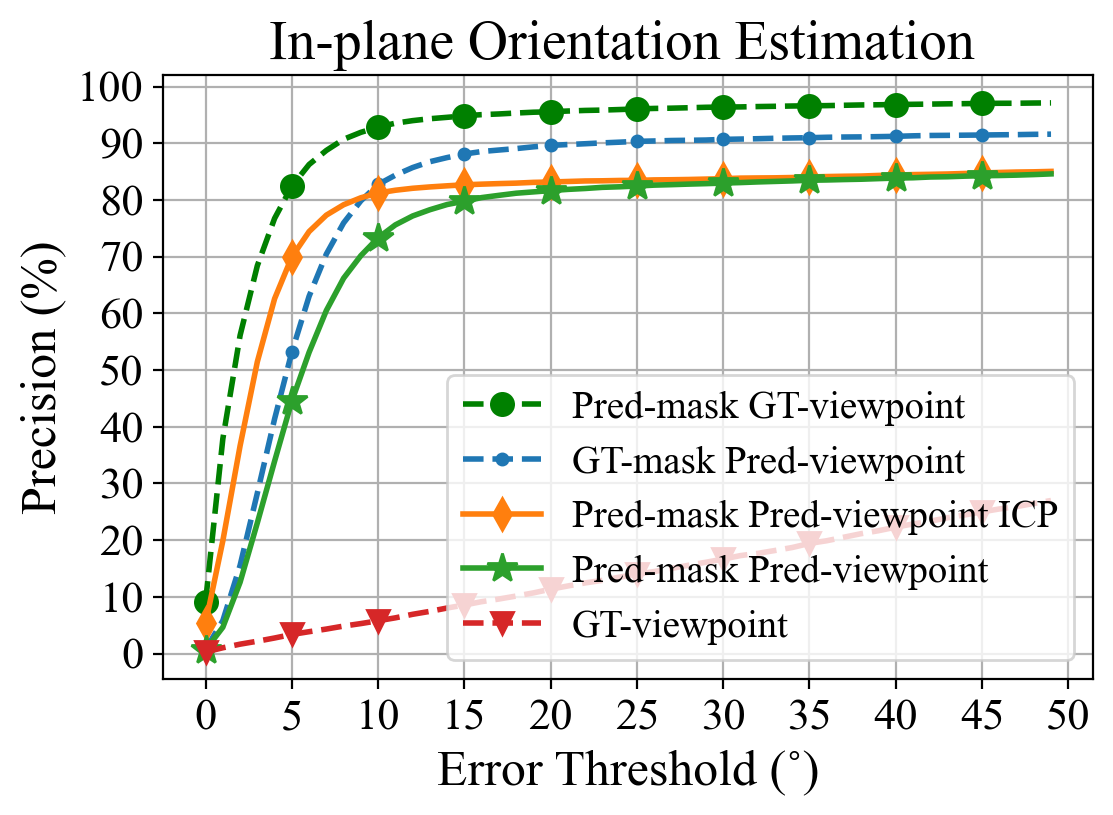}}
  \end{subfigure}
  \begin{subfigure}{0.33\linewidth}
%   {\includegraphics[width=1.\textwidth]{tsl_dve_icp20_T50_1.png}}
  {\includegraphics[width=1.\textwidth]{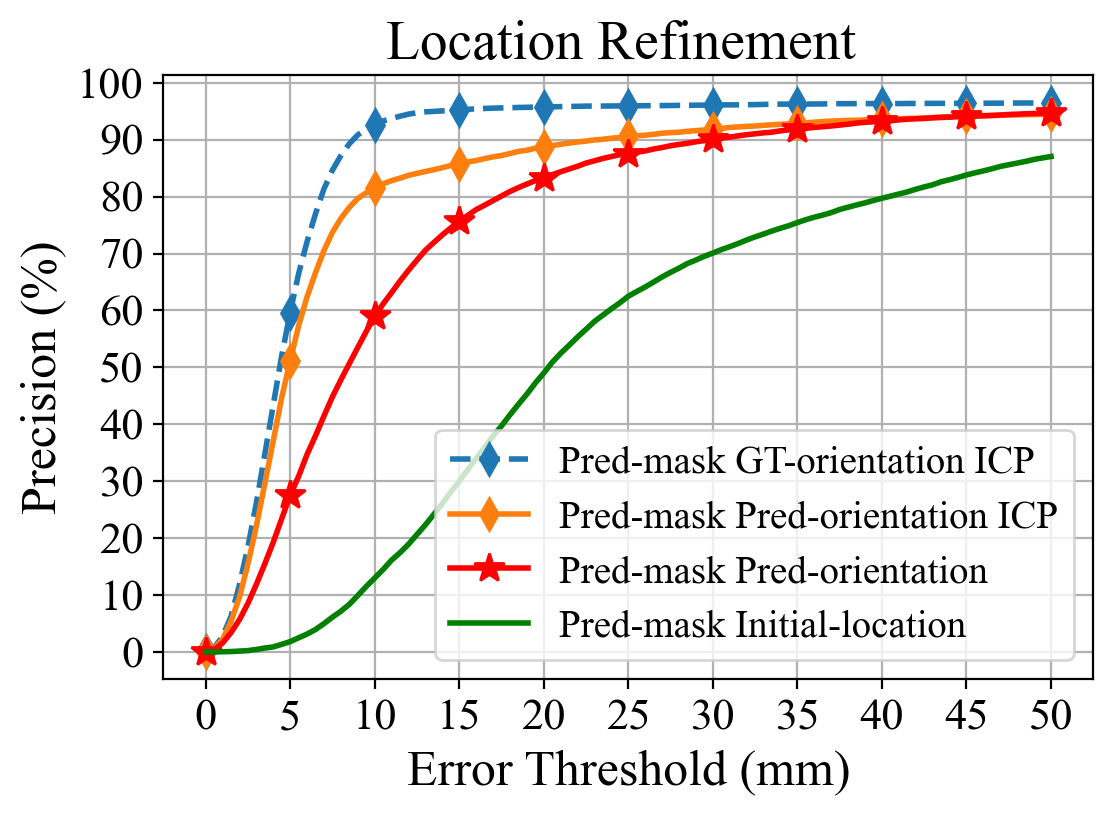}}
  \end{subfigure}
%   \begin{subfigure}{0.35\linewidth}
%   {\includegraphics[width=1.0\linewidth]{seg_mask.png}}
%   \end{subfigure}
  \caption{
  \textbf{Evaluation of OVE6D submodules}. The precision values for different error thresholds for the viewpoint retrieval (left), in-plane orientation regression (middle), and location refinement (right) modules using the LINEMOD dataset.
%   ``GT-mask" and ``Pred-mask'' refer to the ground truth and the Mask-RCNN based object segmentation masks, respectively.  
  ``Real-depth'' and ``Synthetic-depth'' separately refer to real-world depth images and the synthesized depth images (using the ground truth poses).
   ``GT-*" and ``Pred-*'' indicate using the corresponding ground truth and predicted results of *, respectively.  
  %Here, we use only the top-1 viewpoint hypothesis and report the results for viewpoint retrieval, in-plane rotation estimation, and translation estimation on LM. 
    % ``OVE'' in short for the proposed object viewpoint encoder, ``Hist" for histogram, ``GT" for ground-truth, ``Reg" for regression, ``PD" for prediction, ``VP" for viewpoint, ``Ref" for location refinement,  ``init-T" for initial estimate, and ``R" for rotation.
  }
\label{fig:ablation}
\end{figure*}

\paragraph{LINEMOD and Occluded LINEMOD} The results for the LINEMOD (LM) and Occluded LINEMOD (LMO) datasets are reported in tables \ref{table:lm} and \ref{table:lmo}, respectively. All methods, except for OVE6D and LatentFusion\cite{park2019latentfusion}, are specifically trained for the LM dataset. The LatentFusion results are reported only for LM due to heavy occlusions in LMO. In general, the RGBD based methods, trained with real-world and synthetic data, achieve the best performance (\eg 99.7\% recall on LM with FFB6D\cite{he2021ffb6d} and 79.0\% recall on LMO with PVNet\cite{peng2020pvnet}). However, OVE6D obtains competitive results, particularly when compared to the methods trained with purely synthetic data. Without ICP refinement, OVE6D obtains 86.1\% and 56.1\% recall for LM and LMO, respectively. In addition, OVE6D with ICP results in 73\% recall on LMO, which shows that OVE6D is able to generalize to real-world scenes, even in the case of heavy clutter and occlusion. Moreover, compared with LatentFusion\cite{park2019latentfusion}, another universal model trained on ShapeNet, OVE6D obtains better results with a clear margin of 9.3\% (96.4\% vs. 87.1\%) on LM, while relying only depth information in the pose estimation.

\subsection{Additional Experiments}
% We use the LINEMOD dataset to conduct ablation studies on different parameter configurations and verify the effectiveness of the modules in our framework.

\paragraph{Parameter Configuration}
The main parameters in OVE6D are the sampling number of viewpoints (N), the retrieving number of  viewpoint candidates (K), and the number of orientation proposals (P). We examined how these parameter values affect the performance, and observed that the method is relatively stable over a wide range of different settings. We found $N=4000, K=50$, and $P=5$ to be a good trade-off between the accuracy and the efficiency. The detailed results are provided in the supplementary material. 

\paragraph{Viewpoint Retrieval} 
Figure \ref{fig:ablation} (left) illustrates the performance of the viewpoint retrieval module over multiple thresholds using estimated and ground-truth segmentation masks on the LINEMOD dataset. In this experiment, we consider only a single, top scoring, pose hypothesis. 
% For comparison, we add a version which uses a distribution of the depth values instead of the feature embedding from the proposed viewpoint encoder.
We note that already 70\% of the cases are retrieved with higher than $10^{\circ}$ accuracy. In addition, the gap between synthetic and real data is relatively small, indicating good generalization despite of the domain gap.

\paragraph{In-plane Orientation Estimation} 
The results for in-plane rotation module are illustrated in Figure \ref{fig:ablation} (middle). Given viewpoints, retrieved with the predicted masks, we reach up to 73\% precision at the $10^{\circ}$ error threshold with a single forward pass, and further improve it to 80\% with ICP refinement. We also note that by using the ground truth viewpoint, the precision can be further improved over 90\% even without ICP.

\paragraph{Location Refinement} 
Figure \ref{fig:ablation} (right) illustrates the performance of the proposed non-parametric location refinement module. We observe that, at the error tolerance of 10mm, the refinement can improve the precision from the initial estimate 13\% to 60\% and further to 81\% with the ICP refinement. Thus, the proposed refinement module clearly improves the translation estimation with or without ICP refinement.

\paragraph{Inference Time}
The full pose inference with OVE6D requires approximately 50ms per object with Nvidia RTX3090 GPU and AMD 835 Ryzen 3970X CPU. In comparison, LatentFusion requires roughly 20 seconds per object with 100 back propagation iterations. 

\section{Discussion, Limitations and Conclusion}
In this work, we proposed a model called OVE6D for inferring the object 6D pose in a cascaded fashion. The model was trained using a large body of synthetic 3D objects and assessed using three challenging real-world benchmark datasets. The results demonstrate that the model generalizes well to unseen data without needing any parameter optimization, which significantly simplifies the addition of novel objects and enables use cases with thousands of objects. The main limitations of this approach include the requirements for the object 3D mesh model and instance segmentation mask, which may not always be easy to obtain.

\section{Acknowledgement}
This work was supported by the Academy of Finland under project \#327910.

%%%%%%%%% REFERENCES
{\small
\bibliographystyle{ieee_fullname}
\bibliography{OVE6D}
}

\clearpage
\newpage
\appendix
\section{Appendix: supplementary materials}

We provide the supplementary materials for OVE6D in the following.

\begin{figure*}[ht]
\centerline{\includegraphics[width=0.85\textwidth]{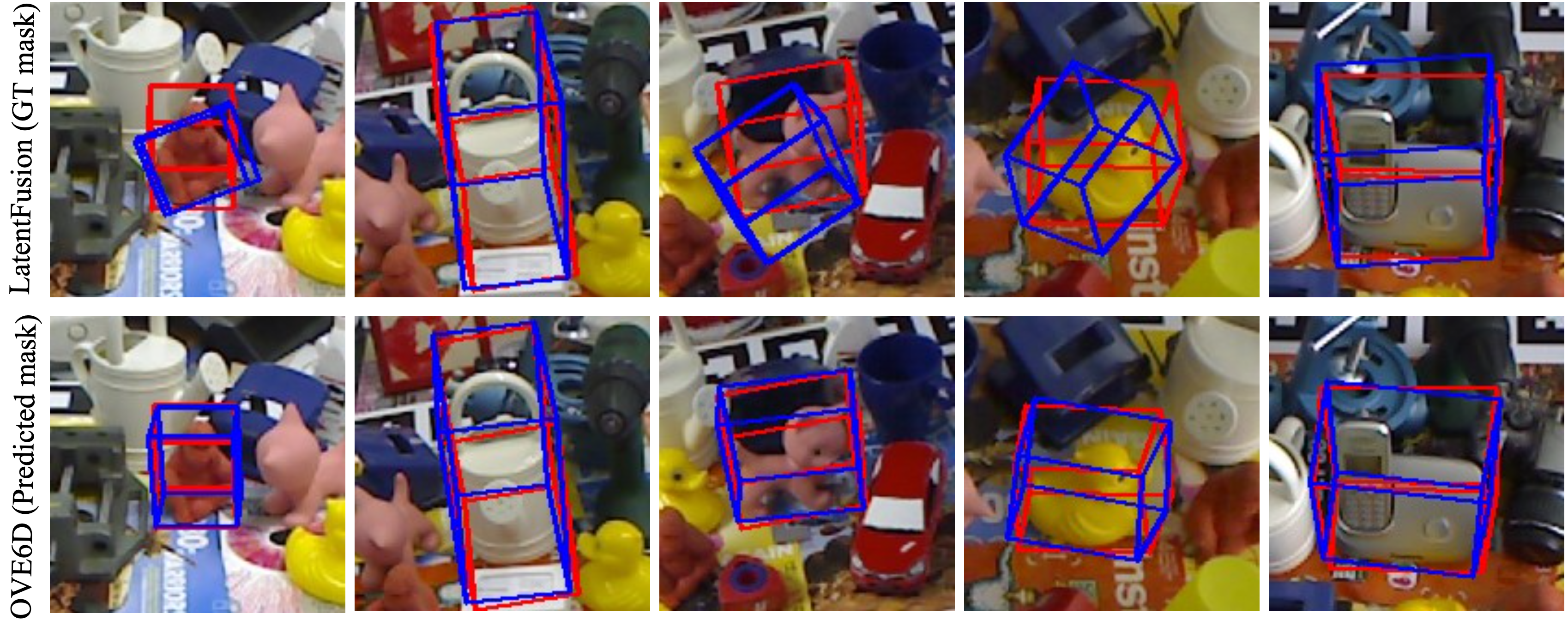}}
\caption{
\textbf{Qualitative evaluation on LineMOD}. We show the qualitative results of LatentFusion\cite{park2019latentfusion} (first row) and OVE6D (second row). {\color{red}{Red}} and {\color{blue}{blue}} 3D bounding boxes indicate the ground truth and the estimated poses, respectively.
}
\label{fig:lm}
\end{figure*}

\subsection{Qualitative examples}
We illustrate qualitative pose estimation examples from the LINEMOD dataset for OVE6D and LatentFusion\cite{park2019latentfusion}  in Figure \ref{fig:lm}. Note that the ground truth segmentation mask is used for LatentFusion by following \cite{park2019latentfusion}, while OVE6D is evaluated using the predicted segmentation mask provided by Mask-RCNN\cite{he2017mask}.  
%We additionally compare some qualitative results between LatentFusion\cite{park2019latentfusion} and OVE6D using the LINEMOD dataset, as shown in Figure \ref{fig:lm}. 

\subsection{Parameter configurations}
The granularity of the discretized out-of-plane rotations (viewpoints) is determined by the parameter $N$. 
We conduct experiments to explore how the ADD(-S) recall is affected by the number of viewpoints $N$, the retrieving number of viewpoint hypothesis $K$, and the number of orientation proposal $P$. 
By increasing the number ($N$) of viewpoints, we reduce the average distance of adjacent viewpoints, which can result in a higher ADD(-S) recall, as shown in Table \ref{table:vp_N}. In addition, retrieving more viewpoint hypotheses ($K$) from the codebook and taking more orientation proposals (P) could increase the probability of obtaining the correct pose for the subsequent stages. The experimental results are presented in Table \ref{table:vp_K} and Table \ref{table:pose_P}.
On the other hand, a finer discretization of the out-of-plane rotation leads to a larger memory footprint (more viewpoints), and more orientation proposals consume a longer verification time. We found $N=4000, K=50$, and $P=5$ to be a good trade-off between the accuracy and the efficiency.

\subsection{Viewpoint codebook construction}
In the main paper, we use the object 3D mesh model to construct the object viewpoint codebook (using synthesized data) to avoid the expensive 6D pose annotation. Nevertheless, the object viewpoint codebook can also be built using real-world training data (with the ground truth 6D object poses). To this end, we conduct additional experiments on the LINEMOD dataset where we build the viewpoint codebook using the real annotated images instead of the mesh model. The experimental results are presented in Table \ref{table:mv_lm} in terms of the average ADD(-S) recall. We can observe a slight gain in the results compared to those obtained with the synthetic data. We attribute this to the alleviation of the domain gap between the object viewpoint codebook and the observed depth images.

\begin{table}[t]
\small
% \scriptsize
\centering
\renewcommand{\arraystretch}{1.2}
\begin{tabular}[0.9\textwidth]{ c | c c c c c} 
\hline\hline
\shortstack{~\\ Sampling Number (N) \\(K = 1, P = 1)} & \shortstack{~\\ 1k \\~} & \shortstack{~\\ 2k \\~} & \shortstack{~\\ 4k \\~} & \shortstack{~\\ 8k \\~} & \shortstack{~\\ 16k \\~} \\
\hline
AAVD(\textdegree) & 6.1 & 4.3 & 3.1 & 2.1 & 1.5 \\
\hline
ADD(-S)(\%) & 74.1 & 75.0 & 75.7 & 75.8 & 76.6 \\
\end{tabular}
\caption{
The average ADD(-S) recalls on the LINEMOD dataset in terms of the varying number of viewpoint sampling. "AAVD" in short for the Average Adjacent Viewpoint Distance.
}
\label{table:vp_N}
\end{table}
\begin{table}[t]
\small
% \scriptsize
\centering
\renewcommand{\arraystretch}{1.2}
\begin{tabular}[0.9\textwidth]{ c | c c c c c} 
\hline\hline
\shortstack{~\\ Retrieving Number (K) \\(N = 4k, P = 1)}  & \shortstack{~\\ 1 \\~} & \shortstack{~\\ 10 \\~} & \shortstack{~\\30\\~} & \shortstack{~\\ 50 \\~} & \shortstack{~\\ 100 \\~} \\
\hline
ADD(-S)(\%) & 75.7 & 80.7 & 81.5 & 81.6 & 81.5 \\
% \hline\hline
\end{tabular}
\caption{
The average ADD(-S) recalls on the LINEMOD dataset in terms of the varying number of viewpoint retrieval.
}
\label{table:vp_K}
\end{table}
\begin{table}[t]
\small
% \scriptsize
\centering
\renewcommand{\arraystretch}{1.2}
\begin{tabular}[0.9\textwidth]{ c | c c c c c} 
\hline\hline
\shortstack{~\\ Proposal Number (P) \\ (N = 4k, K = 50)}  & \shortstack{~\\ 1 \\~} & \shortstack{~\\ 3 \\~} & \shortstack{~\\ 5 \\~} & \shortstack{~\\ 10 \\~} & \shortstack{~\\ 20 \\~} \\
\hline
ADD(-S)(\%) & 81.6 & 85.0 & 86.1 & 87.0 & 87.2\\
\end{tabular}
\caption{
The average ADD(-S) recalls on the LINEMOD dataset in terms of the varying number of orientation proposal.
}
\label{table:pose_P}

\end{table}
\begin{table}[!ht]
\footnotesize
\centering
\renewcommand{\arraystretch}{1.1}
\begin{tabular}{ c | c | c c c c}
\hline
 {\shortstack{~\\Reference\\Data}} & {\shortstack{~\\Method\\~}} & {\shortstack{~\\Input\\~}} & {\shortstack{~\\ICP\\~}}  & {\shortstack{ADD \\ (-S)(\%)}} \\ 
\hline\hline
\multirow{5}{*}{\shortstack{Mutli-View \\ With Pose \\ Annotation}} 
& LatentFusion({GT})\cite{park2019latentfusion} & RGBD & ~ & \textit{87.1} \\
& OVE6D(GT) & D & ~ & \textit{97.0} \\
% & OVE6D(GT) & D & \checkmark & \textit{98.8}\\
& OVE6D(GT) & D & \checkmark & \textit{\bf{99.4}} \\
& OVE6D(MRCNN) & D & ~ & 86.5 \\
% & OVE6D(MRCNN) & D & \checkmark & 92.2\\
& OVE6D(MRCNN) & D & \checkmark & {94.0} \\
\hline
\multirow{4}{*}{\shortstack{Object \\ Mesh \\ Model}} 
& OVE6D(GT) & D & ~ & \textit{96.4}\\%92.2 \\
% & OVE6D(GT) & D & \cmark & \textit{98.3}\\% 95.7 \\
& OVE6D(GT) & D & \cmark & \bf{\textit{98.7}}\\%\bf{97.6} \\
& OVE6D(MRCNN) & D & ~ & 86.1\\%83.6\\%84.1\\ %78.3\\
% & OVE6D(MRCNN) & D & \cmark & 91.4\\%89.0\\%88.2 \\ %84.6\\
& OVE6D(MRCNN) & D & \cmark & {92.4}\\%90.4\\%90.8\\ %88.2\\
\end{tabular}
\caption{Evaluation on \textbf{LINEMOD}. We report the average {ADD(-S)} recall. ICP refinement is performed for all pose proposals before pose selection. MRCNN and {GT} indicate using the masks provided by Mask-RCNN and the ground truth, respectively.
}
\label{table:mv_lm}
\end{table}

 \begin{figure}[t]
\centerline
{\includegraphics[width=0.95\linewidth]{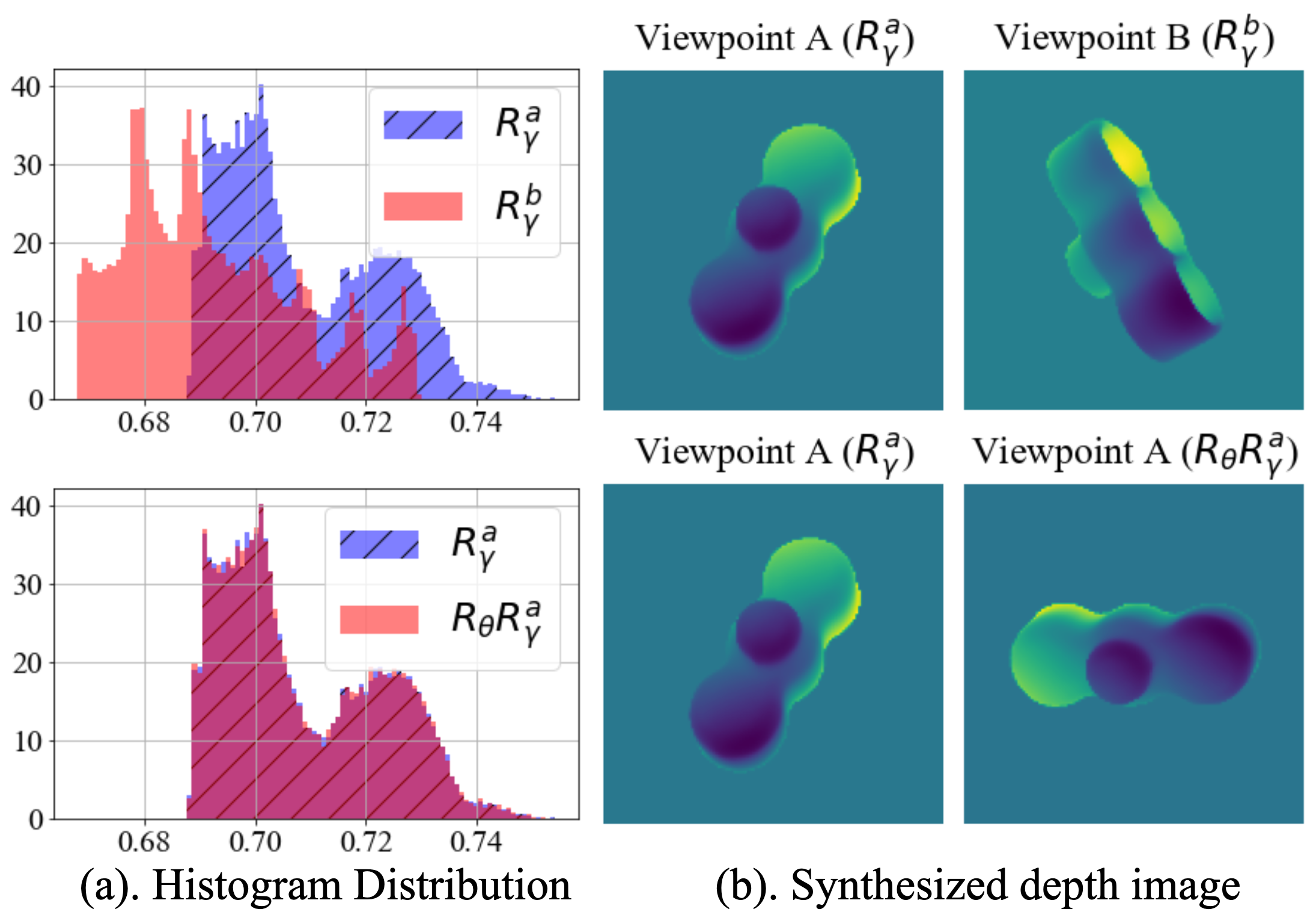}}
\caption{In the first row, we show the histograms of object depth values observed from two different viewpoints $A$ ($\mathbf{R}_{\gamma}^a$) and $B$ ($\mathbf{R}_{\gamma}^b$). In the second row,  we show the histograms of object depth values from the same viewpoint $A$ ($\mathbf{R}_{\gamma}^a$ and $\mathbf{R}_{\theta}\mathbf{R}_{\gamma}^a$). $\mathbf{R}_{\theta}$ is an in-plane rotation around the camera optical axis. We can observe that the (asymmetric) object depth images rendered from different viewpoints result in different distributions. In contrast, the depth images from the same viewpoint but with different in-plane rotations share similar distributions. 
}
\label{fig:hist}
\end{figure}
 \begin{figure*}[ht]
\centerline
{\includegraphics[width=0.85\textwidth]{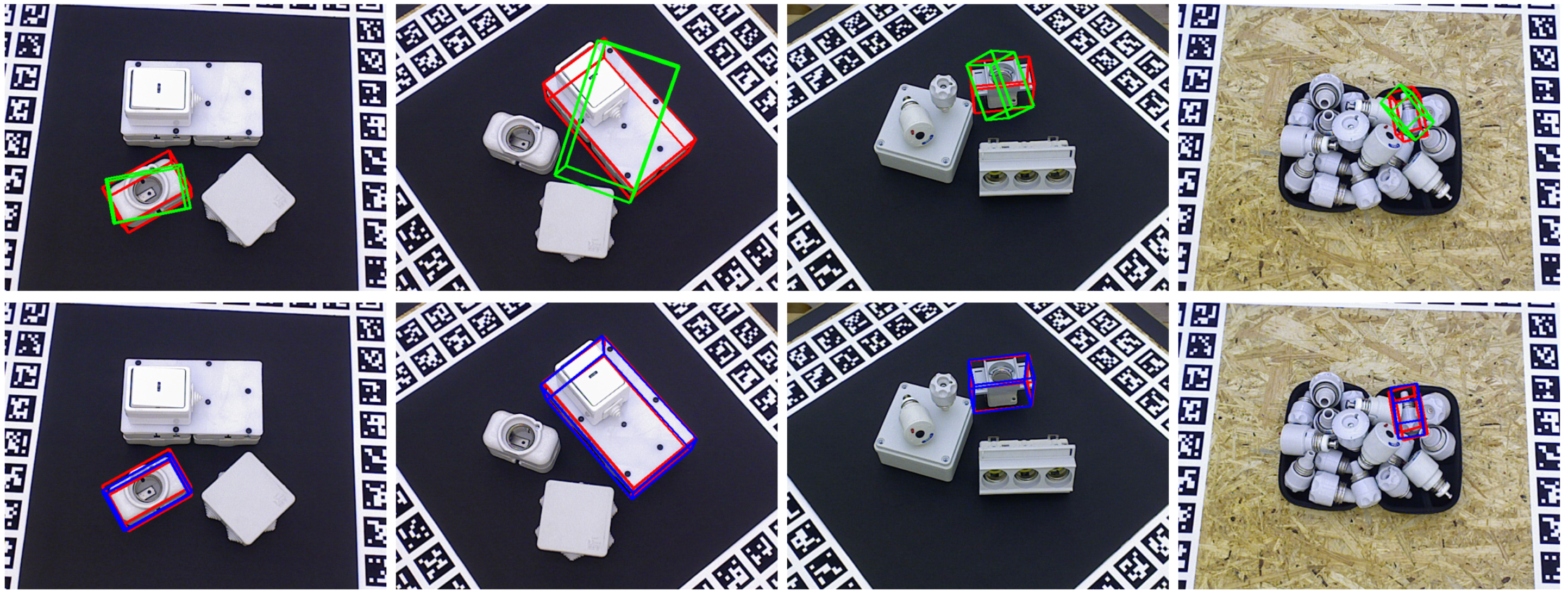}}
\caption{
\textbf{Qualitative evaluation on T-LESS}. In the first row, we show the intermediate results $\mathbf{P}_{temp}$ (without the in-plane orientation regression). In the second row, we show the final complete 6D poses $\mathbf{P}_{final}$ (with the in-plane orientation regression). {\color{red}{Red}}, {\color{green}{green}} and {\color{blue}{blue}} 3D bounding boxes represent the ground truth, the intermediate and the final 6D poses, respectively.
}
\label{fig:tless}
\end{figure*}

\begin{figure*}[t]
\centering
{
\includegraphics[width=.75\linewidth]{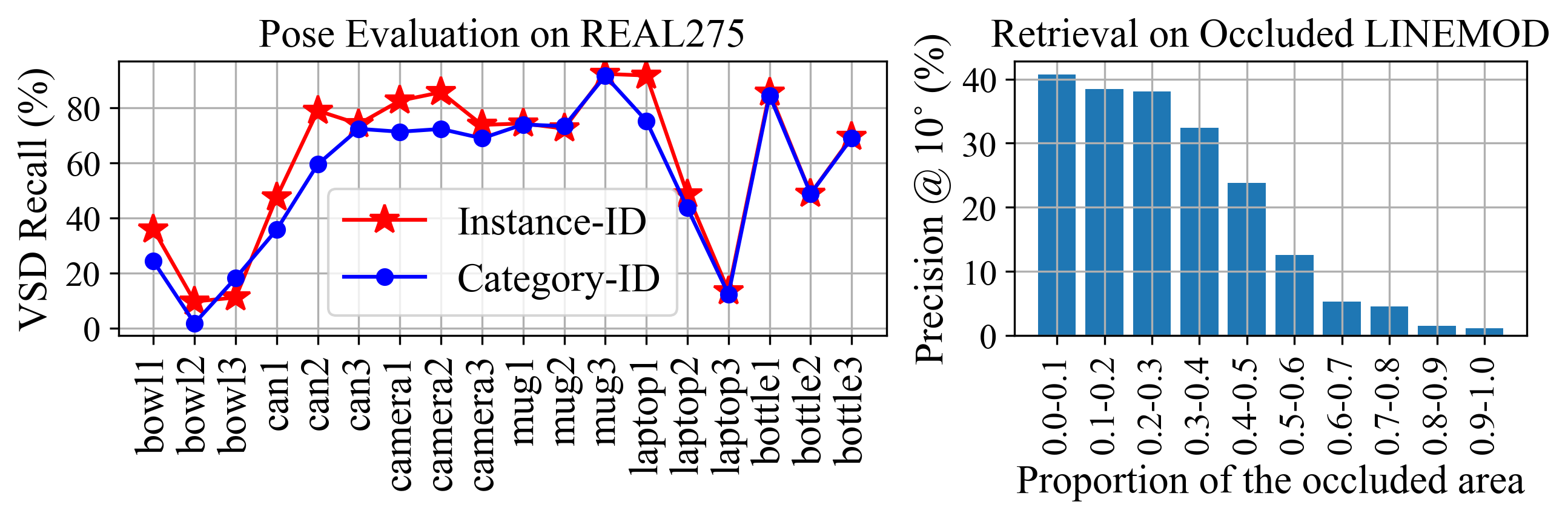}
}
\caption{L: Results for REAL275 (18 objects from 6 categories) using category or instance-level IDs. In category case, we retrieve from all object codebooks belonging to the selected category. R: Viewpoint retrieval results with respect to the occlusion size.}
\label{fig:cat_lmo}
% \vspace{-20pt}
\end{figure*}

\begin{table*}[t]
\small
\centering
\begin{tabular}[0.9\textwidth]{c | c | c} 
% \hline
Technique & Parameter & Description \\
\hline
\shortstack{Rescale} ~ & {0.2 $\sim$ 0.8} & {\shortstack{Downscale the original image with a random ratio and then upscale to the original size.}} \\
\shortstack{LaplaceNoise} & {0.0 $\sim$ 0.01} & {\shortstack{Add the Laplace noise to the downscaled image with a random deviation.}} \\
\shortstack{Cutout} & {0.01 $\sim$ 0.1} & {\shortstack{Cutout rectangular area from the downscaled image with a random area ratio.}} \\
\shortstack{GaussianBlur} & {0.0 $\sim$ 1.5 } & {\shortstack{Apply random Guassian blurring on the downscaled image.}}  \\
\shortstack{RandomOcclusion} & 0.2 &  {\shortstack{Apply a random square or circle occlusion mask on the downscaled image.}}  \\
\end{tabular}
\caption{
Data augmentation techniques and parameters applied on the training data. 
}
\label{table:aug}
\end{table*}

\subsection{Orientation decomposition}
Our method decouples the complete 3D orientation into two components, \ie, the out-of-plane rotation (viewpoint) and the in-plane rotation around the camera optical axis. Here, we provide more details about the factorization. 

The object 3D orientation matrix $\mathbf{R}$ can be factorized into three separate rotations around each axis (x, y and z-axis) with respect to the object coordinate system. \ie
$\mathbf{R} = \mathbf{R}_{z}\mathbf{R}_{y}\mathbf{R}_{x}$, where $\mathbf{R}_{z}$, $\mathbf{R}_{y}$ and $\mathbf{R}_{x}$ are the rotations around the z, y and x axis, respectively. Furthermore, we reformulate the 3D orientation matrix as $\mathbf{R}=\mathbf{R}_{\theta}\mathbf{R}_{\gamma}$, where $\mathbf{R}_{\theta}=\mathbf{R}_z$ is the in-plane rotation around the camera optical axis (z axis) and $\mathbf{R}_{\gamma}=\mathbf{R}_y\mathbf{R}_x$ is the out-of-plane rotation.
In the case of isometric orthographic projection, the histograms of object depth values is mainly determined by the out-of-plane rotation of the object, as illustrated in Figure \ref{fig:hist}. To this end, we uniformly discretize the out-of-plane rotation $\mathbf{R}_{\gamma} \in R^{3\times3}$ as a finite set of object viewpoints $\{\mathbf{R}^{\gamma}_i\}^N_{i=1}$ ($N=4000$ in the main paper) and encode the object viewpoints into latent vectors. These latent viewpoint embeddings are invariant to the in-plane rotation around the camera optical axis. Moreover, the in-plane rotation $\mathbf{R}_{\theta} \in R^{3\times3}$ is formulated as, 
\begin{equation} \label{eq:Rz_matrix}
\mathbf{R}_{\theta} = 
\begin{bmatrix} 
\cos{\theta} &  -\sin{\theta} & 0 \\
 \sin{\theta} & \cos{\theta} & 0 \\
0 & 0 & 1 \\
\end{bmatrix}
\end{equation}
 where $\theta$ is the rotating angle around the camera optical axis. Equivalently, we construct the in-plane rotation matrix $\mathbf{R}_{\theta}$ as,
\begin{equation} \label{eq:Rz_vector}
\mathbf{R}_{\theta} = 
 \begin{bmatrix} 
{\vartheta}_1 &  -{\vartheta}_2 & 0 \\
{\vartheta}_2 & {\vartheta}_1 & 0 \\
0 & 0 & 1 \\
\end{bmatrix}
\end{equation}
where ${\vartheta}_1, {\vartheta}_2$ are scalar values of a unit vector $\Theta\in R^2$ predicted by the in-plane orientation regression network of OVE6D. As presented in Figure \ref{fig:tless}, we show some intermediate results  $\mathbf{P}_{temp}$ without the regressed in-plane rotation and final complete 6D pose results $\mathbf{P}_{final}$ (with the estimated in-plane rotation), \ie,  $\mathbf{P}_{temp}=[\mathbf{R}^{\gamma}_{i}|\mathbf{t}]$ and $\mathbf{P}_{final}=[\mathbf{R}^{\theta}_i\mathbf{R}^{\gamma}_{i}|\mathbf{t}]$,  where $\mathbf{R}^{\gamma}_i\in R^{3 \times 3}$ is the rotation matrix of the retrieved object viewpoint, $\mathbf{t}\in R^{3}$ is the estimated object 3D translation, and $\mathbf{R}^{\theta}_i$ is the estimated in-plane rotation for the retrieved viewpoint.

\subsection{Data augmentation} 
We apply the commonly used training data augmentation techniques to improve the generalization of our model. In particular, we first downscale the synthetic depth image with a random factor and then augment the downscaled depth image (see \cref{table:aug}) before re-scaling it to the original size. The imgaug\cite{imgaug} library is employed to achieve this.

\subsection{Object category-level / instance-level ID} 
We follow the standard practice and use the class labels predicted by the off-the-shelf Mask-RCNN detector as the object IDs. The IDs are used to index the viewpoint codebook and, therefore, a wrong or non-optimal ID could damage the performance as an inadequate codebook would be used in the retrieval. To gain further insight, we evaluated OVE6D using the category-level 6D dataset REAL275 \cite{wang2019normalized}. The results (\cref{fig:cat_lmo} left) show that OVE6D achieves comparable performance using object category IDs instead of the object instance IDs. We believe this is due to the fact that OVE6D is a shape-based method and objects within a category often share similar shapes. 
% Moreover, OVE6D does not \textit{strictly} require RGB image for 6D pose estimation if the segmentation mask can be obtained otherwise. For instance, we could use the Depth Seeding Network \cite{xie2020best} 
% \footnote{Xie \etal The best of both modes: Separately leveraging RGB and depth for unseen object instance segmentation, CoRL, 2020} 
% to predict the object segmentation mask from pure depth data in tabletop environments.

\subsection{Sensitivity to occlusion}
We performed an additional experiment to examine the viewpoint retrieval performance on the Occluded LINEMOD dataset using ground truth segmentation masks in terms of varying percentage of object visibility. The results in Figure \ref{fig:cat_lmo} right indicate that the performance remains almost intact up to 30 \% occlusion and declines smoothly after that. 

\begin{figure*}[ht]
\centering
{
\includegraphics[width=.9\linewidth]{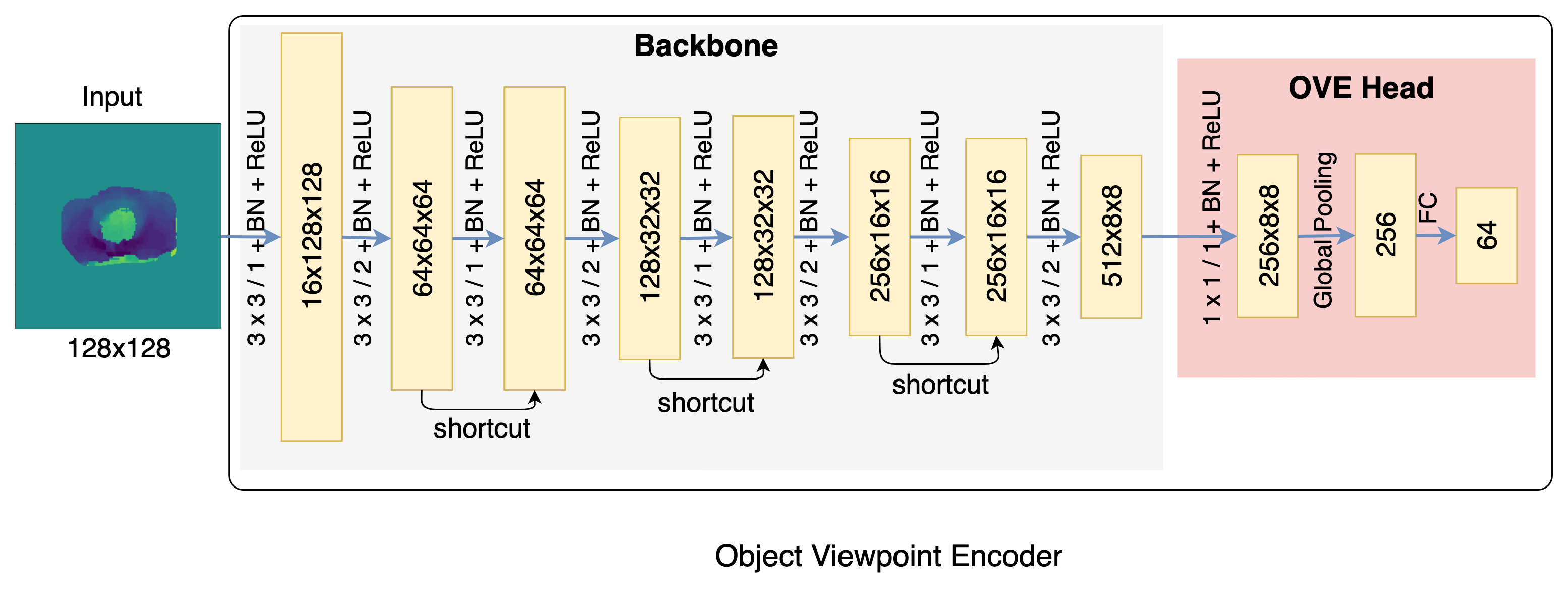}
}
\caption{Network structure of the proposed object viewpoint encoder.}
\label{fig:encoder}
\end{figure*}
\subsection{Structure of the object viewpoint encoder}
Figure \ref{fig:encoder} illustrates the network architecture of the proposed object viewpoint encoder invariant to the in-plane rotation around the camera optical axis. Every convolution layer ($3 \times 3 ~/~ s$ where $s$ denoting stride) is followed by the batch normalization (BN) and ReLU activation layers. Besides, skip connections are added between the feature maps with the same dimensionality.

\subsection{Training details of Mask-RCNN}

We employ Mask-RCNN \cite{he2017mask} from Detectron2 \cite{wu2019detectron2} with the backbone ResNet50-FPN\cite{lin2017feature} to predict the segmenation masks for the objects in the LINEMOD and LINEMOD-Occlusion datasets. We use the physically-based rendered (PBR) images provided by BOP Challenge 2020 \cite{hodan2020bop} to train the network.

Specifically, we apply two steps to finetune the Mask-RCNN to overcome the domain gap between the real and synthetic images. In the first step, we freeze the backbone of Mask-RCNN initialized with the pretrained weights (on MSCOCO dataset \cite{lin2014microsoft}) and train 50k iterations on the training data using the default \textit{WarmupMultiStepLR} learning schedule with the learning rate $lr=0.001$, decayed by 10 at the iteration steps 30k and 40k, respectively. 
In the second step, we unfreeze the backbone and separately train additional 50k iterations for the 13 objects of LINEMOD dataset as well as 50k iterations for the 8 objects of LINEMOD-Occlusion dataset using the \textit{CosineAnnealingLR} learning schedule with the learning rate $lr=0.001$.
% The resulting trained Mask-RCNN reaches an mAP@50 performance of 0.80 for object bounding box detection and 0.77 for segmentation on LINEMOD-Occlusion dataset. Here we do not report the detection results for LINEMOD dataset since each image contains multiple trained objects but only provides a single object ground truth annotation in this dataset.
While for the TLESS dataset we directly employ the segmentation results provided by Multi-Path Encoder \cite{sundermeyer2020multi} for a fair comparison. 
% Sundermeyer \etal \cite{sundermeyer2020multi} trained a Mask-RCNN on 160,000 images generated from T-LESS training set and achieve an mAP@0.5 performance of 0.68 for object bounding box detection and 0.67 for segmentation.

\end{document}